\documentclass[final,1p,times]{elsarticle}

\usepackage{amssymb}
\usepackage{amsmath}

\usepackage[bookmarks=true]{hyperref}
\usepackage{cleveref}
\usepackage{xcolor}
\usepackage{setspace}
\usepackage{amsmath,amsfonts}
\usepackage{algorithmic}
\usepackage{algorithm}
\usepackage{array}
\usepackage[caption=false,font=normalsize,labelfont=sf,textfont=sf]{subfig}
\usepackage{textcomp}
\usepackage{stfloats}
\usepackage{url}
\usepackage{verbatim}
\usepackage{graphicx}
\usepackage{bm}
\usepackage{booktabs}
\usepackage{multirow}
\usepackage{subcaption}
\usepackage{comment}
\usepackage{lineno}
\usepackage{changepage}
\newcommand{\R}[1]{\mathbb{R}^{#1}}
\newcommand{\mX}{\bm{X}}

\newcommand{\mY}{\bm{Y}}
\newcommand{\my}{\bm{y}}
\newcommand{\hY}{\bm{\hat{Y}}}
\newcommand{\hy}{\bm{\hat{y}}}
\newcommand{\revision}[1]{#1} 
\newcommand{\revlast}[1]{#1}

\journal{Ocean Engineering} 

\begin{document}

\begin{frontmatter}



\title{Mix\&Fix-Net: A Dual-Stage Trajectory Prediction Model for AIS and Vision-Derived Vessel Data\tnoteref{fn1}}
\tnotetext[fn1]{This is the accepted manuscript version. The final published version will be available in Ocean Engineering (Elsevier), DOI to be added upon publication.}


\author[label1]{Md Mahmuddun Nabi Murad}
\author[label2, fn2]{Bora San Turgut}
\author[label1]{Yasin Yilmaz}

\affiliation[label1]{
  organization={Department of Electrical Engineering, University of South Florida},
  city={Tampa},
  state={FL},
  postcode={33647},
  country={USA}
}

\affiliation[label2]{
  organization={College of Transportation and Logistics, Istanbul University},
  city={Istanbul},
  country={Turkey}
}


\fntext[fn2]{Bora San Turgut was a visiting scholar at the University of South Florida and was supported by The Scientific and Technological Research Council of Türkiye (TUBITAK) during this work.}

\begin{abstract}
Vessel trajectory prediction is critical for maritime safety and accident prevention. While most existing trajectory prediction models rely on Automatic Identification System (AIS) data due to its precision and availability, small vessels mostly operate without AIS, resulting in a significant monitoring gap. To address this, we propose Mix\&Fix-Net, a dual-stage mixer-based trajectory prediction model designed to handle vessel trajectory time-series data derived from both AIS and (non-AIS) vision data. Our architecture integrates a Primary Trajectory Predictor with a Residual Trajectory Adjuster, enabling more refined trajectory prediction. Additionally, we introduce a new video-based dataset derived from webcam streams, from which vessel trajectories are extracted to represent non-AIS data. 
Extensive evaluations on both AIS and non-AIS datasets across six metrics (mean squared error, mean absolute error, symmetric mean absolute percentage error, final displacement error, Frechet distance, and average Euclidean distance) demonstrate that Mix\&Fix-Net consistently outperforms existing baselines across most metrics and datasets. 
\end{abstract}



\end{frontmatter}


\section{Introduction}
More than three-quarters of global trade occurs using marine transport, and this portion is increasing daily due to the modernization of the marine transport system~\cite{trade}. This increasing demand is also making the system vulnerable to potential hazards. According to~\cite{agcs}, an average of 2,700 maritime accident insurance claims are made to Allianz Global Corporate \& Specialty (AGCS) per year. Additionally, the study in~\cite{unreported} reports that approximately 50\% of maritime accidents go unreported.
These accidents not only cause direct damage to vessels but also result in severe environmental consequences~\cite{environment}. According to~\cite{collision-1, collision-2}, the most frequent accidents are collision-type accidents. Developing an accurate trajectory prediction model is therefore essential to mitigate such accidents and prevent potential hazards.

The traditional trajectory prediction methods utilize the Automatic Identification System (AIS) to locate and estimate the positions of the vessels~\cite{ais}. However, not all vessels have AIS in their system, as AIS is required for vessels of more than or equal to 300 gross tonnage~\cite{SOLAS}. Many smaller vessels, which are not subjected to this requirement, do not have AIS due to the additional installation cost. In this case, it is difficult to monitor the smaller vessels. In addition, the officer on the watch continuously observes the traffic around the ship during the voyage and maneuvers the ship according to situational demands. Nevertheless, even when such vessels are visible to the naked eye, it is still challenging to discern their intentions. Especially in busy maritime traffic areas, where there are many smaller vessels, such as ports, it becomes challenging for the crew or port authority to carefully monitor all surrounding smaller vessels and analyze the risk beforehand. These limitations also highlight the need for a robust trajectory prediction model that remains effective regardless of AIS availability. 

Several prior studies have proposed prediction models, but most rely heavily on AIS data, with only a few exploring alternatives. For instance, the model in~\cite{paper_e} utilizes both the vessel images and AIS data, yet it does not address the previously mentioned challenge of employing an alternative to AIS. The radar image is proposed as an alternative to AIS in~\cite{paper_g}. However, it is not plausible to access the radar image for all types of vessels. Several studies have also explored fusing AIS with other data sources. Ship-radar and AIS data are fused to extract various trajectory characteristics, such as speed, direction, turning rate, and acceleration~\cite{lei2024association}. AIS is combined with video for harbor vessel monitoring~\cite{he2025multi}, and with ship images for maritime computer vision tasks~\cite{paper_h}. AIS is also fused with discretized meteorological features~\cite{huang2022tripleconvtransformer} and multi-source environmental data~\cite{lin2025hdformer} to improve prediction accuracy. However, all of these approaches still depend on AIS as the primary modality. To overcome these limitations and provide a solution applicable to both AIS- and non-AIS-equipped 
systems, we propose Mix\&Fix-Net, a novel dual-stage MLP-mixer-based model designed for vessel trajectory prediction.\par

Vessel trajectory prediction is fundamentally a multivariate time-series prediction task. While RNNs (e.g., LSTM, GRU), TCNs, and Transformers are common in time-series prediction, they often suffer from sequential bottlenecks~\cite{dlinear, tsmixer}. Recent literature shows that MLP-mixer-based architectures can outperform these models by efficiently mixing data in multiple directions~\cite{WPMixer, tsmixer}. Moreover, MLP-Mixer-based models demonstrate strong performance in both long-term and short-term forecasting. Motivated by these advantages, we adopt the MLP-Mixer core in Mix\&Fix-Net to ensure high predictive accuracy in the trajectory prediction task.
Alongside the model design, we introduce a video-based dataset, from which vessel trajectories are extracted, as a non-AIS alternative in scenarios where AIS data are unavailable.

Our contributions can be summarized as follows:
\begin{itemize}
    \item We propose a novel MLP-mixer-based model, Mix\&Fix-Net, that is applicable to both AIS and non-AIS data to predict the trajectory of the vessels. 
    \item We also propose a video-based dataset sourced from an open-access real-time webcam stream for studying vessel trajectory prediction through a practical, easy-to-obtain, non-AIS data modality. 
    \item We provide an in-depth performance analysis for the proposed method by comparing it with existing benchmark methods in terms of six metrics. 
\end{itemize}

\section{Related Work}
\label{sec:relatedworks}
Research on vessel trajectory prediction has become increasingly popular due to its vast applications, such as maritime traffic flow control and accident prevention. Data availability has supported developments in this field. Most of the vessel trajectory prediction studies in the literature are based on AIS data~\cite{wang2025stid, takahashi2024ship, wang2024tptrans, volkova2021predicting}. Due to the easy access to AIS data, numerous studies, using statistical methods \cite{paper_f} and deep learning-based methods, have been performed to predict the trajectory of the vessels. For instance, a transformer-based model, Seaformer, is proposed in \cite{paper_a} to predict the trajectory of the vessel utilizing the AIS data. Similarly, A probabilistic approach for trajectory prediction is proposed in \cite{paper_b}. 

However, AIS latency and data loss negatively affect the accuracy and effectiveness of the vessel intention recognition process~\cite{intersection}. On the other hand, not all vessels are equipped with AIS, and the holistic picture of maritime traffic is not obtained without the localized information of all types of vessels. Therefore, these studies are not complete in this respect. Visual data from widespread cameras on vessels and in ports can greatly contribute to complementing AIS data for vessel trajectory prediction. 

Study in~\cite{paper_c} shows the importance of trajectory prediction in avoiding bridge-ship collisions. They propose a deep learning-based system to avoid bridge-ship collision using multi-channel video data. They integrate the visible-spectrum camera with infrared thermal imaging to address the challenge of trajectory prediction in different illumination conditions. In~\cite{paper_e}, a model based on LSTM and a Graph attention neural network is proposed for vessel trajectory identification and estimation. They also propose a multi-feature-point tracking system utilizing the vessel's movement and orientation in the video. However, the accuracy of trajectory prediction based on video information largely depends on the video quality, while the video quality varies depending on the weather. In~\cite{paper_h}, a novel dataset fusing various images of ships taken in various weather conditions and their corresponding AIS information is proposed. In addition to the video images, multi-sensor information is also utilized to predict the trajectory of the vessels. In~\cite{paper_g}, a multi-task sequence-to-sequence network is proposed for trajectory prediction using the radar image. In addition to radar and visual data, a recent line of research fuses ship-shore VHF voice communication with AIS to enhance maritime intelligent perception. In~\cite{chen2024data}, AIS trajectories are linked with the navigational intentions expressed in VHF speech, associating spoken intentions with vessel tracks via speech recognition and ship-name similarity matching. Additionally, AIS data is integrated with VHF voice feature recognition to predict ship berthing-behavior intentions under noisy inland-waterway conditions~\cite{bai2025prediction}. While these voice-AIS methods add a valuable semantic, intention-level cue, they still depend on AIS as the primary modality and cannot monitor vessels operating without AIS. Our method instead operates on either AIS or vision-derived (non-AIS) trajectories independently, directly addressing this monitoring gap.

With the developments in computer vision technology, the use of cameras as an electronic navigation aid on ships is expected to increase. In parallel, AIS-based trajectory prediction remains critical, as AIS provides precise and reliable vessel position data. Therefore, we propose a novel MLP-mixer-based model that predicts vessel trajectories using historical data derived from either AIS or video sources.

\section{Proposed Method}
\label{proposed_method}
Our objective is to predict the trajectory of the vessels based on their previous positions. We formulate the trajectory prediction problem in the following way. 

\begin{figure*}[!thb]
    \centering
    \includegraphics[width=\textwidth, height=0.5\textheight, keepaspectratio]{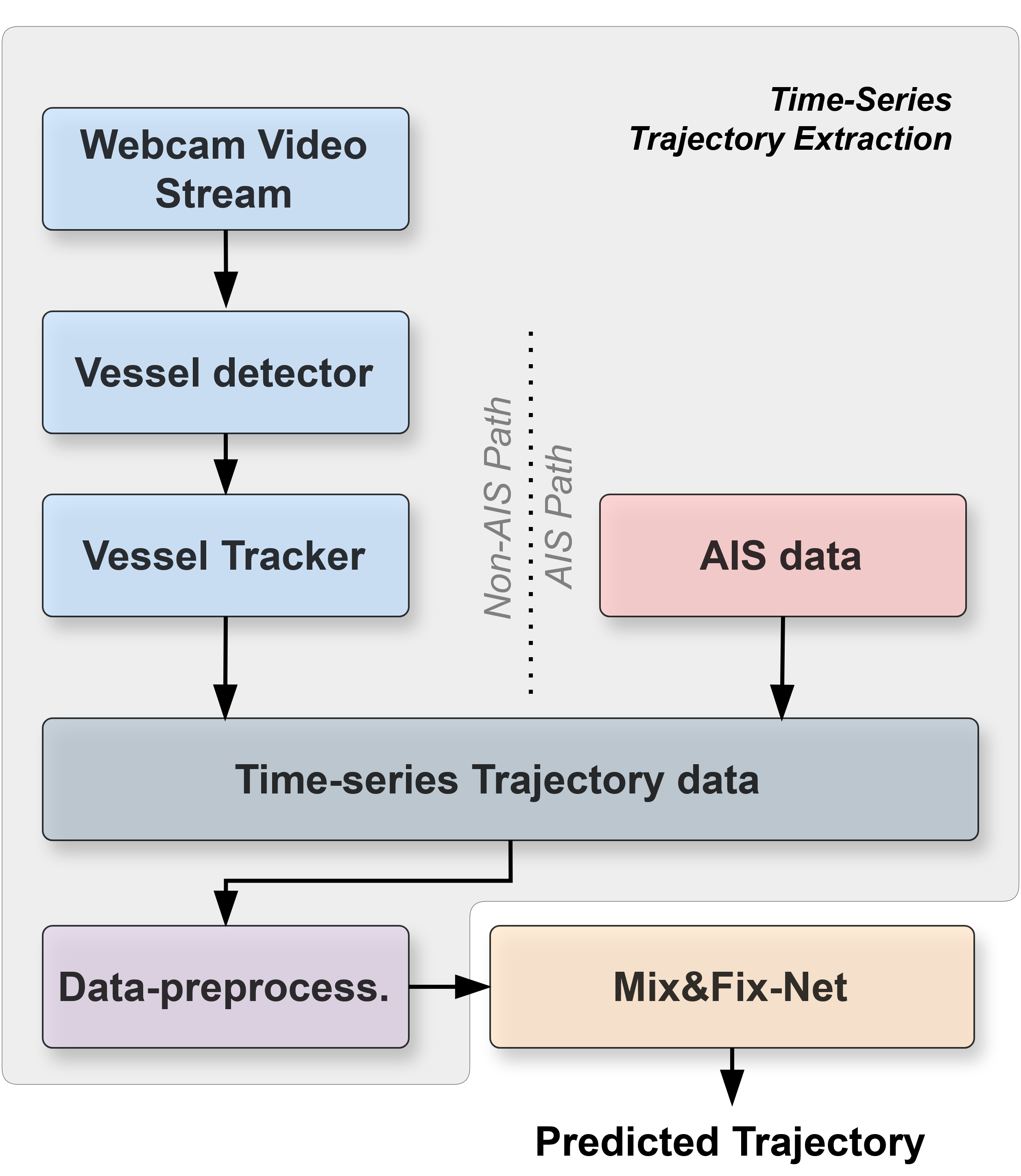}
    \caption{Overall methodological framework of our vessel trajectory prediction pipeline. The framework supports two alternative input modalities: non-AIS (webcam video) processed through a vessel detector and tracker, or AIS data. Mix\&Fix-Net operates on either the non-AIS or the AIS path independently (the two modalities are not fused or used simultaneously). The selected modality produces time-series trajectories that are passed through a data preprocessing stage and fed into Mix\&Fix-Net to predict the trajectory. }
    \label{fig:overall_framework}
    \vspace{-4mm}
\end{figure*}

\begin{figure*}[!thb]
    \centering
    \includegraphics[width=\textwidth, height=0.6\textheight, keepaspectratio]{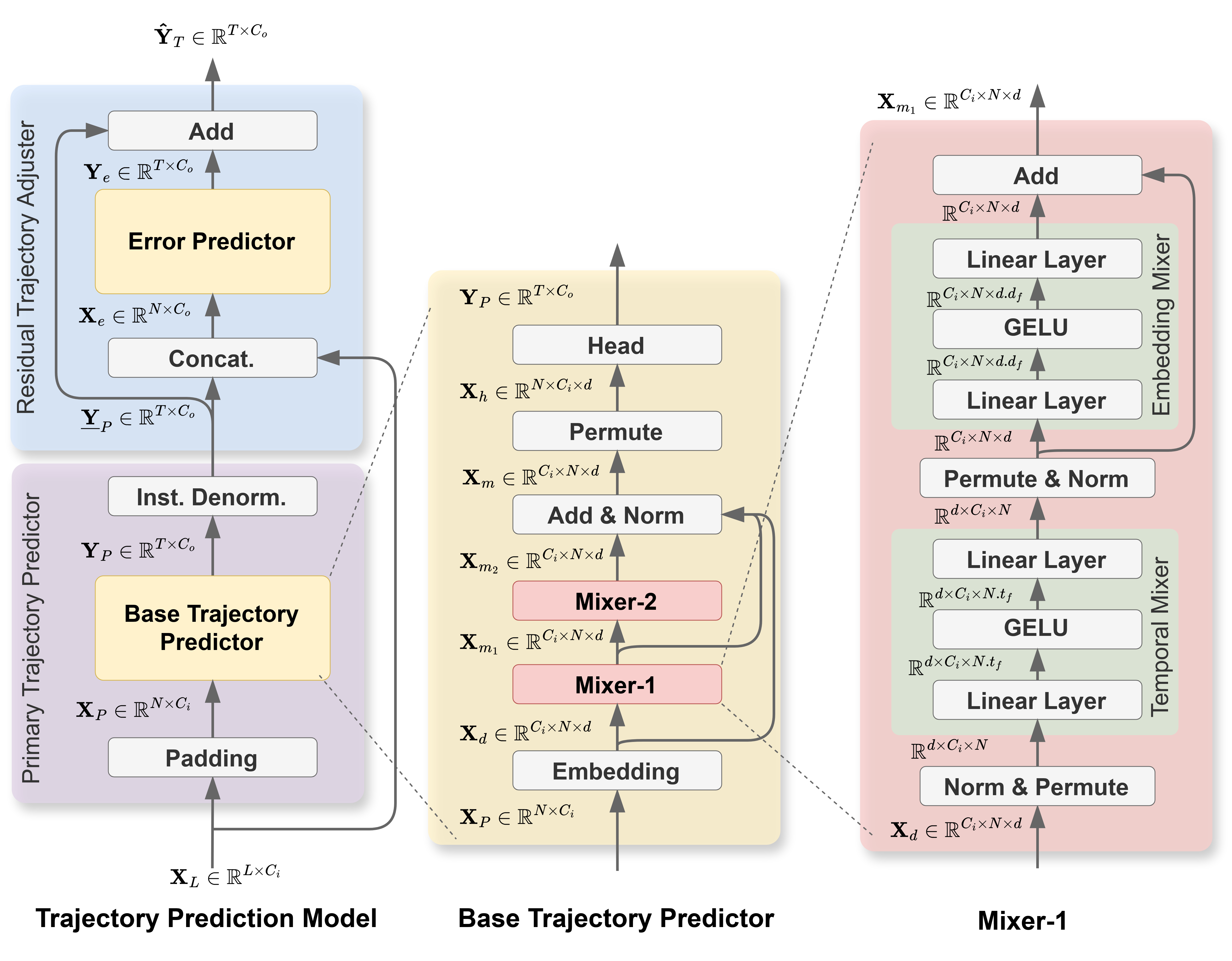}
    \caption{Architecture of our proposed Mix\&Fix-Net, an MLP-mixer-based model, for vessel trajectory prediction.}
    \label{fig:model}
    \vspace{-4mm}
\end{figure*}

Let a multivariate time series $\mX_L \in \R{L \times C_{i}}$ provides the previous $L$ observations of a vessel at time step $t$.
Our goal is to predict the next $T$ positions, denoted by $\hY_T \in \R{T\times C_{o}}$, while the corresponding ground truth is given by $\mY_T \in \R{T\times C_{o}}$. Here, $C_{i}$ denotes the number of input features that include both positional and non-positional information, whereas $C_{o}$ refers to the number of output features that correspond to the positional coordinates. In this work, the time series corresponding to $\mX_L$ can be derived from either AIS or non-AIS modality. The overall framework of our method is shown in Figure~\ref{fig:overall_framework}, which illustrates how time-series trajectories are obtained from the non-AIS data, while the AIS data is directly used as time-series. The detailed architecture of our proposed trajectory prediction model, Mix\&Fix-Net, is shown in Figure~\ref{fig:model}. 

\subsection{Overall Modular Architecture of Mix\textnormal{\&}Fix-Net}
\label{overall_model_architecture}
Mix\&Fix-Net is a dual-stage MLP-mixer-based model. In the first stage, it predicts the coarse trajectory of the vessel using the Primary Trajectory Predictor module, while refining the prediction in the second stage with the Residual Trajectory Adjuster module.

\medskip
\noindent\textbf{Primary Trajectory Predictor:} The Primary Trajectory Predictor comprises zero-padding of the input, the Base Trajectory Predictor module, and the instance denormalization. Initially, we append $T$ zero-padding entries to the end of the input $\mX_L \in \R{L \times C_{i}}$ to obtain $\mX_P \in \R{N \times C_{i}}$, where $N = L + T$. These padded values serve as placeholders for additional information during model processing, which is particularly beneficial when $T > L$. Subsequently, the Base Trajectory Predictor processes $\mX_P$ to generate the initial coarse trajectory prediction $\mY_P \in \R{T \times C_o}$. $\mY_P$ then undergoes instance denormalization~\cite{revin}, resulting final coarse trajectory prediction $\underline{\mY_P} \in \R{T \times C_o}$. Instance denormalization facilitates the Base Trajectory Predictor’s ability to handle input distribution shifts, with statistical parameters (mean and standard deviation) computed from $\mX_L$.

\medskip
\noindent\textbf{Residual Trajectory Adjuster:} The Residual Trajectory Adjuster refines the coarse prediction by estimating the residual error. We concatenate the positional features of $\mX_L$ with $\underline{\mY_P}$, forming $\mX_e \in \R{N \times C_o}$, where $N = L + T$. The Error Predictor then predicts the error $\mY_e \in \R{T \times C_o}$. The final trajectory prediction $\hY_T \in \R{T \times C_o}$ is computed as,
\begin{align}
\hY_T = \underline{\mY_P} + \mY_e
\end{align}

This dual-stage design with coarse trajectory prediction followed by residual refinement improves prediction accuracy by effectively modeling both the primary trajectory and fine-grained adjustments. 

\subsubsection{Base Trajectory Predictor and Error Predictor Modules}
\label{Base_Route_Predictor_and_Error_Predictor_Modules_appendix}

The architectures of the Base Trajectory Predictor and Error Predictor modules are identical, differing only in the number of input features. The Base Trajectory Predictor module receives $C_{i}$ features as input, whereas the Error Predictor module receives $C_o$ features. Aside from this distinction, both modules consist of an embedding layer, multiple mixer modules, and a head module. Below, we demonstrate the sequential data processing of the Base Trajectory Predictor.\par
Initially, the input $\mX_P \in \R{N \times C_{i}}$ of the Base Trajectory Predictor is embedded to $\mX_d \in \R{C_{i} \times N \times d}$ employing an embedding layer, where $d$ denotes the embedding dimension for each channel. Subsequently, $\mX_d$ is processed by an MLP-Mixer module (Mixer-1) and outputs $\mX_{m_1}$. Then $\mX_{m_1}$ is processed by the second MLP-Mixer module (Mixer-2), resulting in $\mX_{m_2}$. We then add $\mX_{m_2}$ to  $\mX_{m_1}$ and $\mX_d$ through residual connections and perform layer normalization $\mathcal{LN}(.)$: 
\begin{align}
    \mX_m = \mathcal{LN}(\mX_{m_2} + \mX_{m_1} + \mX_d) \in \R{C_{i} \times N \times d}.
\end{align}
Later, we permute $\mX_m$, and obtain $\mX_h \in \R{N \times C_{i} \times d}$. Finally, $\mX_h$ undergoes the head module to obtain the initial coarse prediction $\mY_p \in \R{T \times C_o}$.\par
In the following subsections, we provide a detailed description of the embedding layer, Mixer module, and Head module.

\subsubsection{Embedding Layer} Embedding is performed to encode the data into a higher-dimensional space using a linear layer. Here, the embedding layer transforms $\mX_P \in \R{N \times C_i}$ into a higher dimensional representation $\mX_d \in\R{N \times C_{i}\cdot d}$. The embedding layer $\mathcal{L}_{emb}$ is defined as,
\begin{align}
    \mathcal{L}_{emb}: \R{1 \times C_i} \longrightarrow \R{1 \times (C_{i}\cdot d)},
\end{align}
where $d$ is a hyperparameter. After the linear transformation, each embedding vector is reshaped and permuted as follows,
\begin{align}
    \R{1 \times C_i} \xrightarrow{\text{embed}} \R{1 \times (C_{i} \cdot d)} \xrightarrow{\text{reshape}} \R{1 \times C_{i} \times d} \xrightarrow{\text{permute}} \R{C_{i} \times 1 \times d}
\end{align}
to obtain a $d$-dimensional  representation for each channel.

\subsubsection{Mixer Module}
Both the Base Trajectory Predictor and Error Predictor modules employ two sequential mixer modules (Mixer-1 and Mixer-2) to facilitate multidirectional data mixing. Each mixer module also contains two distinct mixer blocks: the temporal mixer and the embedding mixer. The architecture of the temporal mixer and embedding mixer is similar to the token-mixing MLP described in \cite{c:mlpmixervision}. Below, we describe the sequential data processing in the first mixer module (Mixer-1).\par
Initially, the input $\mX_d \in \R{C_i \times N \times d}$ is normalized using layer-normalization, and permuted to obtain the dimension $\R{d \times C_i \times N}$. This rearrangement ensures that the last dimension of the data corresponds to $N$, along which temporal mixing is subsequently applied.\par

\noindent\textbf{Temporal Mixer:} The temporal mixer mixes data along the temporal dimension $N$. It comprises two linear layers connected by a non-linear GELU activation. The dimensional transformation of the data in the temporal mixer is shown as,
\begin{align}
    \R{\cdots \times N} \xrightarrow{\text{Linear}} \R{\cdots \times (N\cdot t_f)} \xrightarrow{\text{GELU}} \R{\cdots \times (N\cdot t_f)} \xrightarrow{\text{Linear}} \R{\cdots \times N}
\end{align}
Here, the first linear layer mixes information along the $N$ dimension and expands it by a factor of $t_f$, a hyperparameter. The final linear layer re-transforms the expanded dimension $(N\cdot t_f)$ to the original $N$ dimension.\par
The output of the temporal mixer, with dimension $\R{d \times C_i \times N}$, is permuted and normalized using layer normalization to obtain $\R{C_i \times N \times d}$. The permutation operation ensures that the last dimension of the data corresponds to $d$, along which the embedding mixing is performed subsequently.\par

\noindent\textbf{Embedding Mixer:} The Embedding mixer mixes data along the embedding dimension $d$. Similar to the temporal mixer, the embedding mixer also comprises two linear layers connected by a non-linear GELU activation. Dimensional transformation of the data in the embedding mixer is shown as,
\begin{align}
    \R{\cdots \times d} \xrightarrow{\text{Linear}} \R{\cdots \times (d\cdot d_f)} \xrightarrow{\text{GELU}} \R{\cdots \times (d\cdot d_f)} \xrightarrow{\text{Linear}} \R{\cdots \times d}
\end{align}
Here, the initial linear layer mixes the information along the $d$ dimension and expands it by a factor of $d_f$, which is a hyperparameter. The final linear layer projects the expanded dimension $(d\cdot d_f)$ back to the original dimension~$d$.\par
To obtain the final output $\mX_{m_1} \in \R{C_i \times N \times d}$ from  Mixer~-1, the output of the embedding mixer module is added to its input, ensuring residual learning.

\subsubsection{Head Module}
\label{head}
The input to the Head module is denoted as $\mX_h \in \R{N \times C_i \times d}$, where each temporal instance, with shape $\R{1 \times C_i \times d}$, contains an embedding vector of length $d$ for each channel. The objective of the head module is to transform each temporal representation from $\R{1 \times C_i \times d}$ into $\R{1 \times C_i}$. Therefore, we first flatten the last two dimensions of $\R{1 \times C_i \times d}$, and subsequently perform a linear transformation using a linear layer: 
\begin{align}
    \R{1 \times C_i \times d} \xrightarrow{\text{Flatten}} \R{1 \times (C_i \cdot d)} \xrightarrow{\text{Linear}} \R{1 \times C_i}.
\end{align}
By performing these operations on the $N$ temporal data points of $\mX_h \in \R{N \times C_i \times d}$, we obtain data with dimension $\R{N \times C_i}$. Since our ultimate goal is to predict the $T$ temporal instances with $C_o$ features, we select the last $T$ temporal data instances from $N$ and retain the last $C_o$ features from $C_i$, resulting in the final output $\mY_p \in \R{T \times C_o}$.

\begin{figure}[!htb]
  \centering
  \subfloat[VVR]{%
    \includegraphics[width=0.33\textwidth]{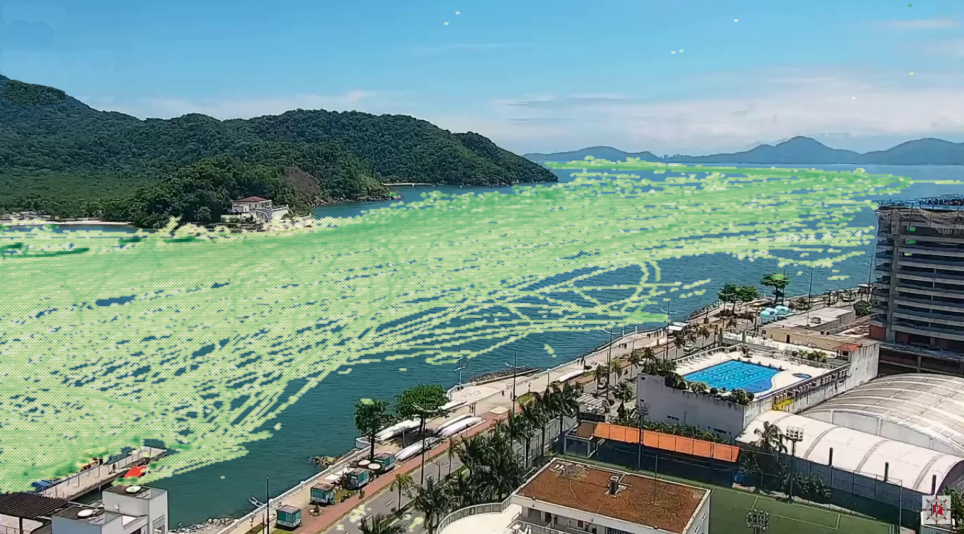}%
    \label{fig:mapA}}
  \hfill
  \subfloat[M3]{%
    \includegraphics[width=0.33\textwidth]{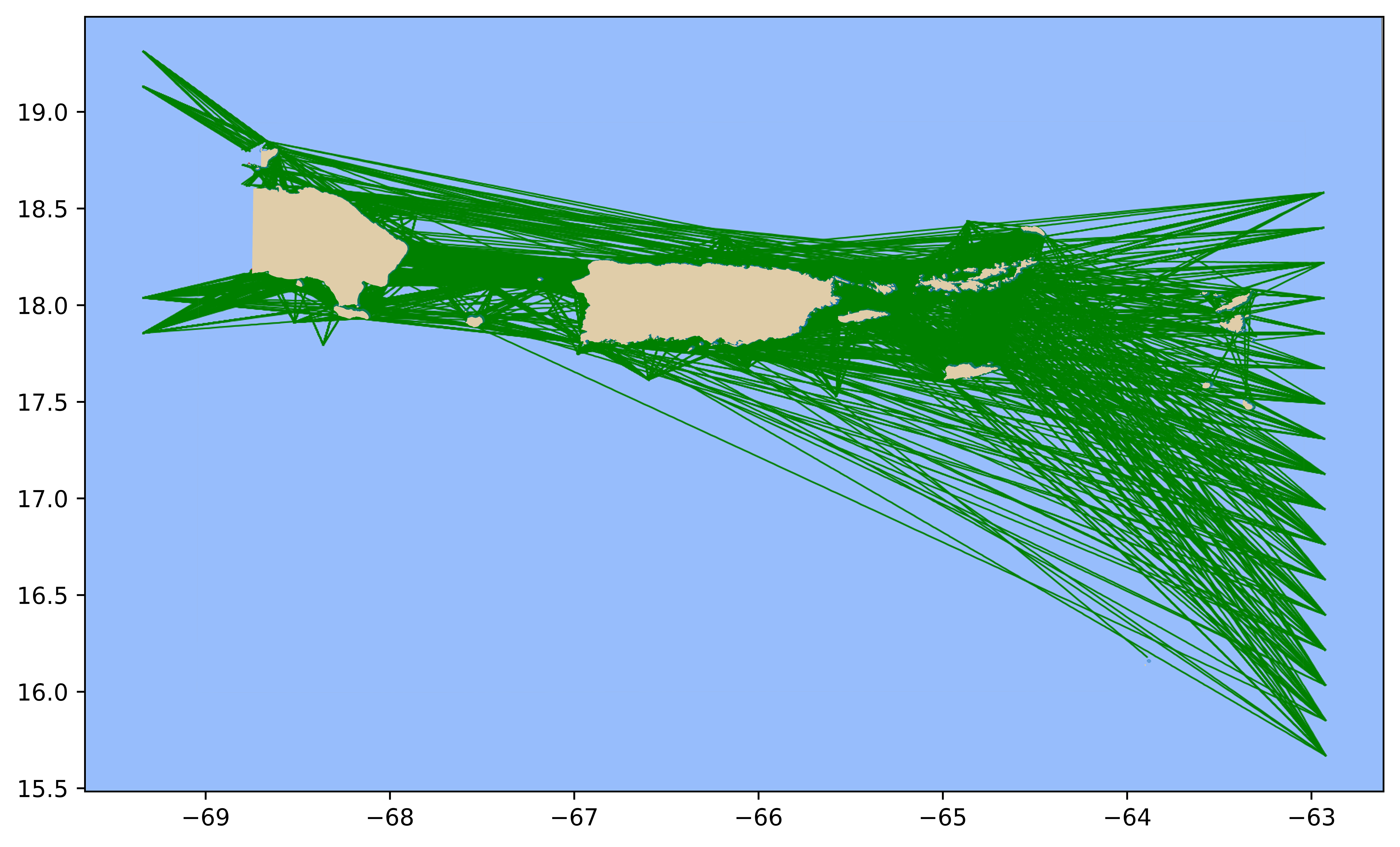}%
    \label{fig:mapB}}
  \hfill
  \subfloat[TampaBay]{%
    \includegraphics[width=0.33\textwidth]{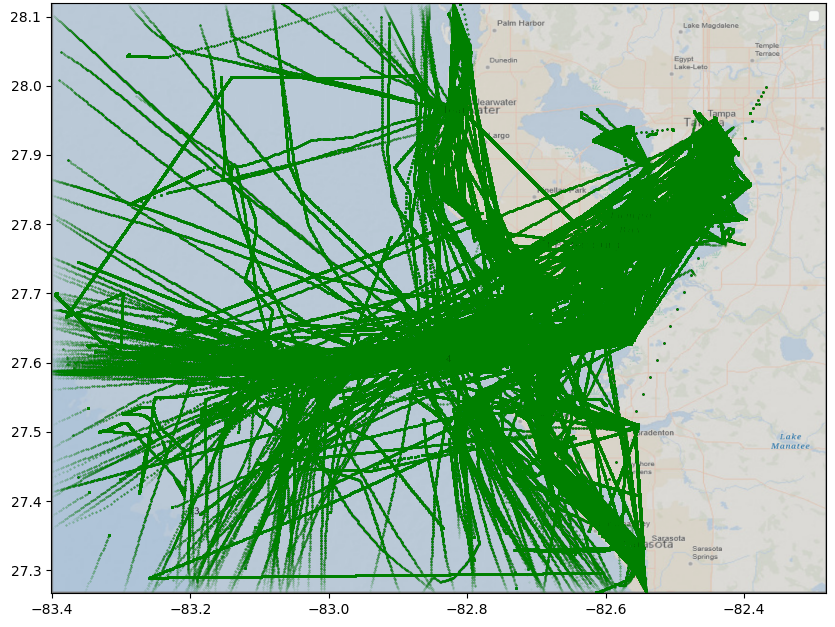}%
    \label{fig:mapC}}
  \caption{Vessel trajectory maps of the study areas for the VVR, M3, and TampaBay datasets.}
  \label{fig:two-maps}
  
\end{figure}

\begin{figure*}[!htb]
  \centering
  \subfloat{\includegraphics[width=0.33\textwidth]{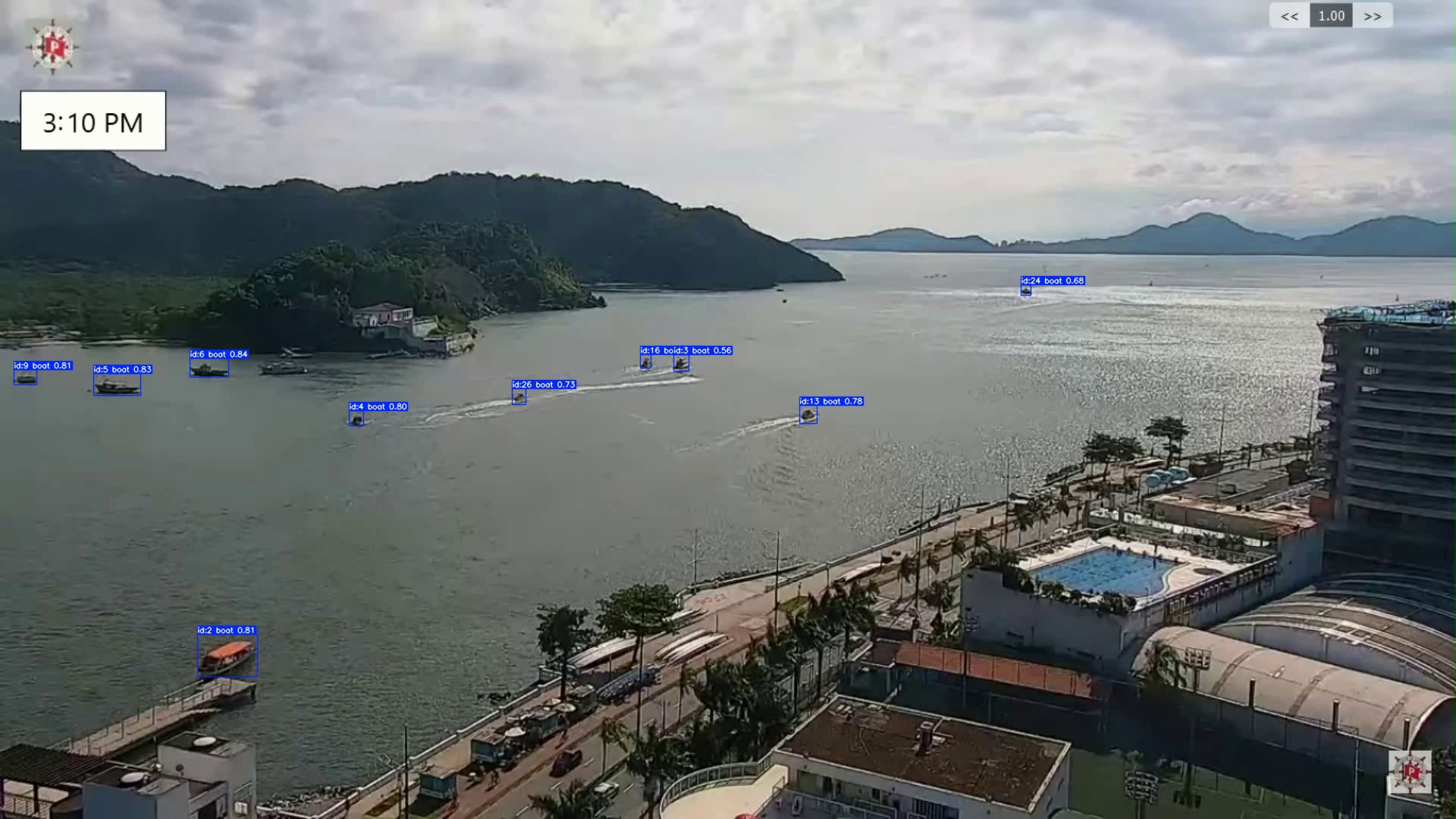}}\hfill
  \subfloat{\includegraphics[width=0.33\textwidth]{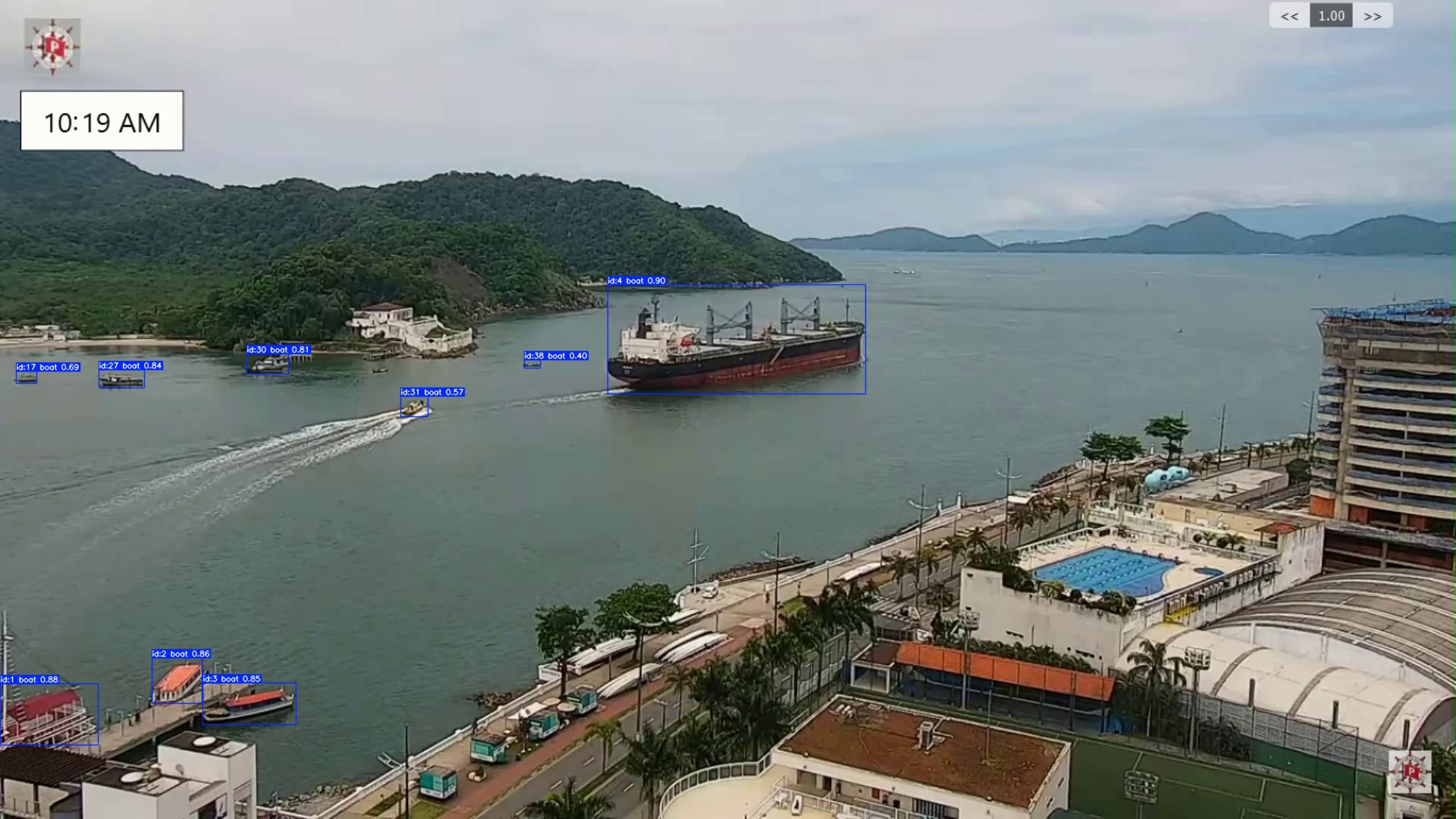}} \hfill
  \subfloat{\includegraphics[width=0.33\textwidth]{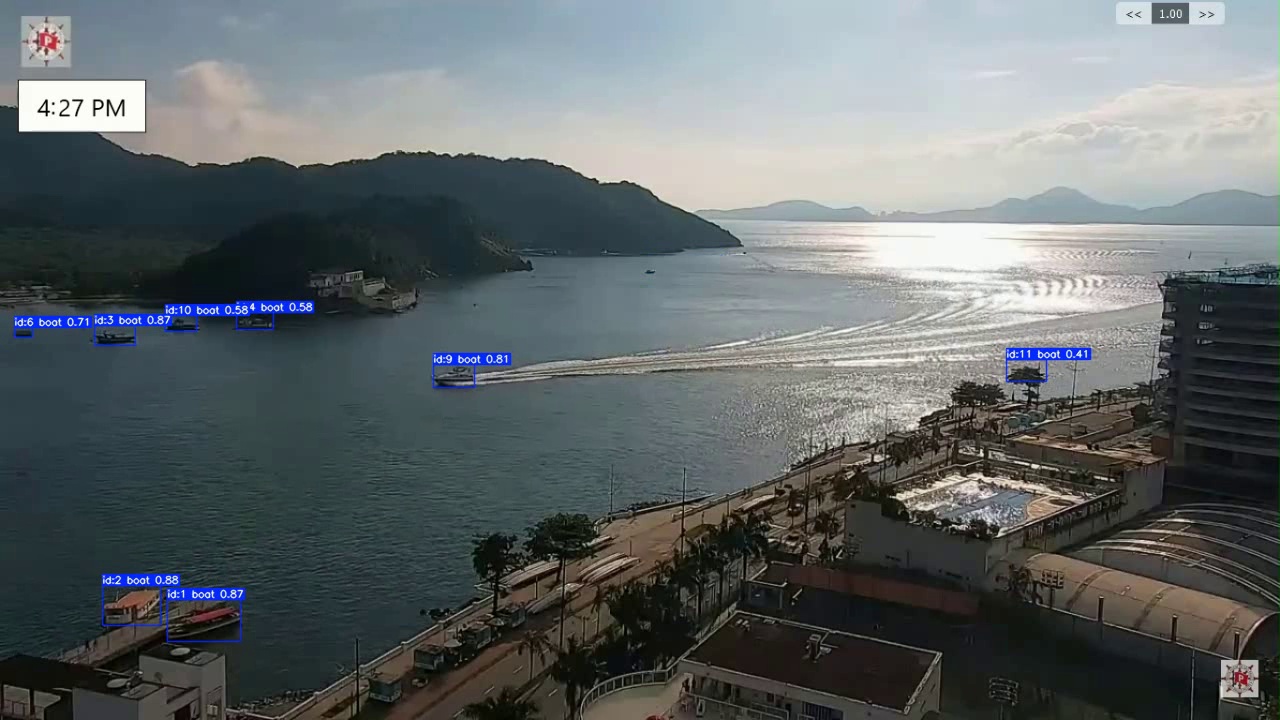}}\hfill
  \caption{Vessel tracking in the VVR dataset in different weather conditions. Each vessel is assigned a new vessel-ID.}
  \label{fig:vessel_tracking_images}
\end{figure*}

\subsection{Data Collection}
\label{dataset_generation_and_acquisition}
To evaluate the trajectory prediction performance of our model, we employ three benchmark datasets: one non-AIS dataset and two AIS-based datasets. For the non-AIS dataset, we introduce the Video-based Vessel Route (VVR) dataset~\footnote{The VVR dataset is publicly available at \url{https://github.com/Secure-and-Intelligent-Systems-Lab/Route-Prediction-Model}.}, in which vessel trajectories are extracted from video frames and represented as time-series data. For the AIS data, we utilize the Managing Maritime Movements (M3) Data Challenge dataset~\cite{m3hackathon2024}\footnote{The M3 dataset is restricted and requires authority approval.}. Since M3 is not publicly available, we supplement our evaluation with a second, publicly available AIS dataset obtained from~\cite{MarineCadastre2024}. Figure~\ref{fig:two-maps} illustrates the vessel trajectory mappings for these datasets within their respective study areas. The following sections describe the collection and acquisition processes for these datasets in detail.

\subsubsection{Collection of non-AIS data}
\label{generation_of_non_ais_data}
Our proposed model predicts vessel trajectories based on data provided in a time-series format. Consequently, the model is applicable to any domain where sequential time-series data can be obtained. While AIS data is inherently sequential and can be directly adopted for the prediction task, our model is also evaluated using non-AIS data. Specifically, we employ non-AIS time series that are generated from videos. Unlike AIS data, video-based data requires stable recordings to capture vessel motion. Publicly available surveillance streams often fail to meet this requirement due to limited coverage or moving cameras. To overcome this, we selected a real-time webcam feed from the Canal do Porto de Santos, an estuary in Santos, Brazil, which continuously records vessel traffic from a fixed viewpoint \cite{video_stream}. The dataset consists of six distinct video recordings spanning three days, totaling 24 hours of video at a resolution of 1366×768 and a frame rate of 30 fps. The videos capture vessels of varying sizes, ranging from small boats to large cargo ships, and span diverse environmental conditions throughout the day. This variability enhances the diversity of the dataset, which can be used to assess algorithm robustness. To transform video footage into time-series data, we utilize the following procedure:
\begin{itemize}
    \item \textbf{Vessel Detection:} First, a pre-trained YOLO11 model is fine-tuned using manually annotated vessel images from the first two videos. This tuned model is then used to detect vessels within the video frames. The model generates a bounding box for each detected vessel, and the bounding box center is used as the vessel's position.
    \item \textbf{Tracking:} Using the fine-tuned detector, we first detect vessels on the remaining four videos and track them using the ByteTrack algorithm. Tracker assigns a unique vessel ID to each tracked trajectory. Examples of vessel tracking are shown in Figure \ref{fig:vessel_tracking_images}. This tracking process links positions across video frames to specific timestamps.
\end{itemize}
As a result, we obtain a comprehensive time-series dataset containing vessel IDs, timestamps, and corresponding $xy$ coordinates (center of bounding boxes). The values of $x$ and $y$ range from 0 to 1, where (0, 0) corresponds to the top-left corner and (1, 1) corresponds to the bottom-right corner on the video frame. The details of the Video-based Vessel Route (VVR) dataset are shown in Table \ref{tab:video_dataset_details}. Detailed implementations of the YOLO model and ByteTracking are provided in ~\ref{vessel_detection_and_tracking_imp_details}.

\subsubsection{Acquisition of AIS datasets}
\label{acquisition_of_ais_datasets}

\noindent {\bf{Managing Maritime Movements (M3) dataset:}} The M3 dataset is a synthetic time series dataset that provides latitude and longitude coordinates for 10,000 vessels in an area with latitude from 14.817 to 22.300 and longitude from -69.375 to -62.930. The initial features include profile ID, vessel ID, timestamp, latitude, and longitude. The profile ID ranges from 1 to 10 and corresponds to the portfolio that governs how trajectories are generated. The time frame of the dataset is from January 14, 2022, to January 16, 2022. The M3 dataset contains two sets: the training set, comprising trajectories for 8,000 vessels, and the test set, reserved for data challenge submissions, which includes trajectories for 2,000 vessels. We utilize the training set of the data challenge to train and validate our proposed model.\par

\noindent {\bf{TampaBay dataset:}} We collect a real-world AIS dataset covering the period from January 01, 2024, to January 20, 2024, via the marinecadastre.gov website. The raw data is filtered to focus on the Tampa Bay area, specifically within the geographic bounds of $27.27^\circ$ to $28.12^\circ$ Latitude and $-83.40^\circ$ to $-82.28^\circ$ Longitude. The dataset includes key features such as Maritime Mobile Service Identity (MMSI), timestamp, vessel type, latitude, and longitude, where MMSI and vessel type serve as functional equivalents to the Vessel ID and Profile ID features used in the M3 dataset. In total, the TampaBay dataset comprises 981 unique vessel IDs and 26 distinct vessel types.

\subsection{Dataset Preprocessing}
\label{dataset_preprocessing}
We perform a series of preprocessing steps on the time-series trajectories of the TampaBay, M3, and VVR datasets before training our trajectory prediction model. These steps include removing duplicate time instances, interpolating missing values, and performing feature engineering. Below, we demonstrate these steps in a sequential order.

\subsubsection{Removing duplicate time instances}
We identify several repeated data points with identical timestamps in the trajectory of the M3 and TampaBay datasets. To resolve this issue, we retain only the first data point and remove all subsequent repeated data points along the trajectory. This step is not applied to the VVR dataset because it contains no repeated data points.

\subsubsection{Interpolating missing values}
We observe that the tracker model fails to detect the vessels on certain video frames. As a result, the vessels' trajectories lack uniform time sampling. Similarly, the M3 and TampaBay time series datasets also miss several temporal data instances. Therefore, we apply linear interpolation to estimate and fill the missing data points. After interpolating the missing values, the sampling periods are $1/30$ second for the VVR dataset, and 1 minute for the M3 and TampaBay datasets. Since the sampling rate of the VVR dataset is excessively fine-grained, predicting a trajectory of a certain length requires a lot of data points, which makes the computation expensive. To mitigate this, we downsample the VVR dataset by a factor of 10, resulting in a final sampling period of $1/3$ second.

\subsubsection{Adding additional features}
Initially, the M3 and TampaBay include profile ID, vessel ID, timestamp, latitude, and longitude, while the VVR dataset includes the same features with the exception of the profile ID. We transform the timestamp into day ($D$), hour ($H$), and minute ($M$) values. We then compute the sine and cosine values of $D$, $H$, and $M$ by the following formulas:
\begin{align}
M_{\sin} &= \sin\!\left(\frac{2\pi M}{60}\right) &\quad
M_{\cos} &= \cos\!\left(\frac{2\pi M}{60}\right) \label{eq:minute}\\
H_{\sin} &= \sin\!\left(\frac{2\pi H}{24}\right) &\quad
H_{\cos} &= \cos\!\left(\frac{2\pi H}{24}\right) \label{eq:hour}\\
D_{\sin} &= \sin\!\left(\frac{2\pi D}{7}\right)  &\quad
D_{\cos} &= \cos\!\left(\frac{2\pi D}{7}\right)  \label{eq:day}
\end{align}
We also incorporate displacement $D_{disp}$ as a feature, defined as the distance traveled between consecutive time steps. Consequently, the final features for the VVR are $M_{sin}$, $M_{cos}$, $H_{sin}$, $H_{cos}$, $D_{sin}$, $D_{cos}$, $D_{disp}$, $x$-coordinate (longitude) and $y$-coordinate (latitude), while the M3 and TampaBay additionally include the profile ID. Therefore, the VVR dataset consists of 9 features, whereas the M3 and TampaBay consist of 10 features.

\section{Experiments}
This section presents the experimental framework, including dataset splitting, dataset normalization, evaluation metrics, baselines, training configurations, results, and ablation studies.

\subsection{Dataset Splitting}
\label{data_splitting}
In trajectory prediction literature, two primary splitting strategies exist: (1) extracting sequences via a rolling window before distributing them into sets~\cite{yang2024vessel, li2024vessel}, and (2) partitioning raw data by vessel ID or time prior to sequence extraction~\cite{evmides2025vessel, slaughter2025vessel}. The former approach carries a risk of data leakage because overlapping segments of the same trajectory may appear in both training and testing sets. Some studies use hybrid variants of these methods to handle varying testing conditions~\cite{guo2025vessel}. In this work, we evaluate our model using both experimental settings to ensure a robust analysis of generalization:

\noindent {\bf{Setting-1:}} To strictly prevent data leakage and evaluate the model's ability to generalize to unseen vessels, we first partition the dataset into training, validation, and test sets with a ratio of 3:1:1, based on vessel IDs. For the TampaBay dataset, we maintain this ratio within each vessel profile and for each day; if a profile lacks sufficient vessel IDs to satisfy this ratio, it is excluded. For the M3 dataset, the ratio is maintained within each profile, while for the VVR dataset, it is applied across all vessel IDs. Following the split, we extract observation-prediction pairs from each vessel's trajectory using a one-step stride rolling window. To focus on active movement, we exclude pairs where the standard deviation of coordinates in the observation window is less than $0.001$. This setting ensures the model is evaluated on entirely new trajectories not encountered during training.

\noindent {\bf{\revision{Setting-2:}}} We first apply the same rolling window and zero-motion filtering to each trajectory independently. These filtered pairs are then randomly distributed into training, validation, and testing sets with a 3:1:1 ratio. \par

In both settings, we set the observation and prediction window lengths for all datasets to 60 and 120, respectively, aligning with the criteria of the M3 data challenge. This choice also considers the trade-off with the minimum required trajectory length $(L+T)$, as many trajectories are shorter. Additionally, reducing the window size negatively affects the model performance. A detailed justification for the selected window lengths is provided in~\ref{hyper_sensiti_appendix}. Total sequence counts for both configurations are reported in Table~\ref{tab:data_sequences}.

\begin{table}[!htb]
\centering
\small
\caption{Details of the VVR dataset. The dataset consists of six video recordings. Vessels are manually annotated in two videos to train the vessel detector, while the remaining four videos are used to convert vessel trajectories into time series representations. Times reported in the table correspond to local time.}
\label{tab:video_dataset_details}
\begin{tabular}{@{}ccccc@{}}
\toprule
S/N & Recording Date & Start time & Duration & Purpose \\ \midrule
0 & 2/12/2025 & 13:02:00 & 6 Hours & Detector training \\
1 & 11/19/2024 & 8:01:00 & 4 Hours & Detector training \\
2 & 11/19/2024 & 12:17:00 & 4 Hours & Trajectory Prediction \\
3 & 11/19/2024 & 16:26:00 & 38 Minutes & Trajectory Prediction \\
4 & 11/20/2024 & 8:12:00 & 4 Hours & Trajectory Prediction \\
5 & 2/12/2025 & 7:50:00 & 5 Hours & Trajectory Prediction \\ \bottomrule
\end{tabular}
\end{table}

\begin{table}[!htb]
\centering
\small
\caption{Number of observation-prediction sequence pairs.}
\label{tab:data_sequences}
\begin{tabular}{@{}r|ccc|ccc@{}} \toprule
\multicolumn{1}{c}{} & \multicolumn{3}{c|}{\revision{Setting-1}} & \multicolumn{3}{c}{\revision{Setting-2}} \\ \cmidrule{2-7}
\multicolumn{1}{c}{} & M3 & TampaBay & VVR & M3 & TampaBay & VVR \\ \midrule
Train & 367039 & 133740 & 119312 & 369845 & 200719 & 117421 \\
Validation & 127752 & 49651 & 37361 & 123282 & 66940 & 39140 \\
Test & 121618 & 60817 & 39029 & 123282 & 67046 & 39141 \\ \midrule
Total & 616409 & 244208 & 195702 & 616409 & 334705 & 195702 \\ \bottomrule
\end{tabular}
\end{table}

\subsection{Dataset Normalization}
\label{dataset_normalization}

We normalize both the training and test sets using the mean and standard deviation (z-normalization) derived from the training set. To be more precise, the input to our model $\mX_L$ denotes the normalized input. Since the predicted output $\hY_T$ is compared against the normalized ground truth $\mY_T$ during model training, $\hY_T$ also represents the normalized prediction. However, when evaluating model performance, we denormalize both $\mY_T$ and $\hY_T$ to their original scale, denoted as $\mY \in \R{T \times C_o}$ and $\hY \in \R{T \times C_o}$, respectively. Here, $C_o$ indicates the number of the output features, which corresponds to either $xy$-coordinates or the latitude-longitude position of the vessel.\par

While AIS and VVR datasets utilize different coordinate systems (geographic vs. image-plane), they are unified through z-normalization to focus on motion dynamics. A detailed discussion on the semantic consistency and coordinate-agnostic nature of our model is provided in~\ref{semantic_consistency_of_coordinate_systems}.

\subsection{Evaluation Metrics}
\label{evaluation_metrics}
To comprehensively evaluate the performance of our model, we employ six evaluation metrics, as outlined in \cite{metrics}: MSE, MAE, SMAPE, FDE, FD, and AED. MSE (Mean Squared Error) measures the average squared distance between predicted and true positions at each time step, whereas MAE (Mean Absolute Error) captures the average absolute difference between them. SMAPE (Symmetric Mean Absolute Percentage Error) quantifies the relative difference between the predicted and true positions. FDE (Final Displacement Error) measures the Euclidean distance between the predicted and true final positions, while FD (Frechet Distance) captures the maximum distance between the predicted and true trajectories. Finally, AED (Average Euclidean Distance) computes the mean Euclidean distance between predicted and true positions across the predicted time steps.
We compute the six metrics for each predicted sequence by the following formulas,

\begin{align}
\text{MSE} & = \frac{1}{T} \sum_{i=1}^{T} \|\hy_i - \my_i\|_2^2 &\quad
\text{MAE} = \frac{1}{T} \sum_{i=1}^{T} \|\hy_i - \my_i\|_1 \nonumber \\
\text{SMAPE} & = \frac{1}{T} \sum_{i=1}^{T} \frac{\|\hy_i - \my_i\|_1}{\big(\|\hy_i\|_1 + \|\my_i\|_1\big)/2} &\quad
\text{FDE} = \|\hy_T - \my_T\|_2 \nonumber \\
\text{FD} & = \max_{i \in [1,T]} \|\hy_i - \my_i\|_2  &\quad
\text{AED} = \frac{1}{T} \sum_{i=1}^{T} \|\hy_i - \my_i\|_2 \nonumber
\end{align}

where, $\|.\|_2$ and $\|.\|_1$ refer to L2 norm and L1 norm, respectively, $\hy_i \in \R{1 \times 2}$ denotes the $i$-th predicted position in the predicted trajectory $\hY$, and $\my_i \in \R{1 \times 2}$ denotes the corresponding true position in the ground-truth trajectory $\mY$. The final value for each metric is obtained by averaging again over the total number of sequences.

\subsection{Baselines}
The number of publicly available codes for existing trajectory prediction models is limited. Most existing models are adapted from widely used architectures that have demonstrated strong performance in sequential forecasting tasks, as trajectory prediction is closely aligned with time series forecasting. Recently, Transformer, MLP-mixer, and RNN-based models have achieved notable success in time series forecasting. Therefore, as baselines for our model, we select four variants that have shown superior performance in forecasting: P\_sLSTM~\cite{P_sLSTM}, PatchTST~\cite{c:patchtst}, DLinear~\cite{dlinear}, and WPMixer~\cite{WPMixer}. In addition, we include a domain-specific trajectory prediction model, FusionRNN~\cite{slaughter2025vessel}.

Among these, P\_sLSTM is an sLSTM-based model that incorporates patching and was originally proposed for time series forecasting. To enhance the effectiveness of Transformer models for forecasting, PatchTST introduces patching into the Transformer architecture, achieving superior performance compared to other Transformer-based variants. DLinear is a simple yet effective MLP-based model that challenges the Transformer-based models by questioning their performance for the forecasting task. Recently, MLP-Mixer-based models have often outperformed Transformer variants. WPMixer is one such model, extending the MLP-mixer framework by utilizing wavelet decomposition in combination with the patching technique. FusionRNN~\cite{slaughter2025vessel} is a fusion-based RNN architecture specifically proposed for AIS-based vessel trajectory prediction, which demonstrated state-of-the-art performance on its corresponding real-world AIS datasets.

\subsection{Training and Hyperparameters}
\label{train_hyperparameters}
During training, model optimization is carried out with the Adam optimizer and MSE loss function. All experiments on the VVR and TampaBay datasets are conducted using a single NVIDIA GeForce RTX 4090 GPU. Given that the M3 dataset is approximately three times larger than the VVR dataset, training is accelerated by utilizing two NVIDIA A100 GPUs. All experiments on the VVR and TampaBay datasets used a batch size of 128, while those on the M3 dataset used a batch size of 512.

For hyperparameter tuning, we utilize Optuna, a hyperparameter tuning framework.
The search space includes: embedding dimension $d \in \{16, 32, 64, 128, 256\}$, embedding mixer factor $d_f \in \{1, 3, 5, 7\}$, temporal mixer factor $t_f \in \{1, 3, 5, 7\}$, and learning rate ranging from 0.01 to 0.00001. The maximum number of epochs is fixed at 50 with an early-stopping patience of 10 for all experiments. The detailed hyperparameter settings are reported in Table~\ref{tab:hyperparams_tables} in the Appendix.

\begin{table}[!htb]
\centering
\small
\setlength{\tabcolsep}{3pt}
\caption{Performance comparison under data split \revision{setting-1}. For each model, the best results are reported. The best-performing model is highlighted in bold, while the second-best is underlined.}
\label{tab:result_setting_2}
\begin{tabular}{@{}cccccccc@{}}
\toprule
 &  & Mix\&Fix-Net & WPMixer & P\_sLSTM & PatchTST & Dlinear & FusionRNN \\ \midrule
\multirow{6}{*}{\rotatebox[origin=c]{90}{VVR}} & MSE & \textbf{5.22E-04} & 7.76E-04 & \underline{6.71E-04} & 8.18E-04 & 9.83E-04 & 9.22E-04\\
 & MAE & \textbf{1.58E-02} & \underline{1.61E-02} & \textbf{1.58E-02} & 1.66E-02 & 2.05E-02 & 2.35E-02 \\
 & SMA. & \textbf{4.60E-02} & 5.22E-02 & \underline{4.78E-02} & 5.44E-02 & 6.31E-02 & 7.01E-02 \\
 & FDE & \textbf{2.69E-02} & 3.05E-02 & \underline{2.95E-02} & 3.27E-02 & 3.99E-02 & 3.60E-02 \\
 & FD & \underline{3.21E-02} & 3.29E-02 & \textbf{3.13E-02} & 3.36E-02 & 4.06E-02 & 3.96E-02 \\
 & AED & \textbf{1.28E-02} & \underline{1.30E-02} & \underline{1.30E-02} & 1.35E-02 & 1.63E-02 & 1.88E-02 \\ \cmidrule(l){1-8}
\multirow{6}{*}{\rotatebox[origin=c]{90}{M3}} & MSE & \textbf{1.38E-03} & \underline{2.95E-03} & 4.19E-03 & 3.16E-03 & 3.69E-03 & 5.60E-03 \\
 & MAE & \textbf{2.60E-02} & \underline{3.11E-02} & 4.24E-02 & 3.20E-02 & 3.51E-02 & 4.02E-02 \\
 & SMA. & 2.89E-03 & \underline{1.83E-03} & 1.07E-02 & \textbf{1.79E-03} & 3.45E-03 & 2.03E-02 \\
 & FDE & \textbf{4.28E-02} & 6.36E-02 & 8.15E-02 & 6.80E-02 & 7.23E-02 & \underline{5.41E-02} \\
 & FD & \textbf{5.08E-02} & 6.44E-02 & 8.42E-02 & 6.86E-02 & 7.41E-02 & \underline{6.03E-02} \\
 & AED & \textbf{2.07E-02} & \underline{2.46E-02} & 3.47E-02 & 2.55E-02 & 2.79E-02 & 3.17E-02 \\ \cmidrule(l){1-8}
\multirow{6}{*}{\rotatebox[origin=c]{90}{TampaBay}} & MSE & \textbf{2.69E-03} & 3.19E-03 & \underline{2.92E-03} & 3.28E-03 & 3.18E-03 & 3.66E-03 \\
 & MAE & \underline{3.88E-02} & \textbf{3.87E-02} & 4.02E-02 & 4.08E-02 & 4.27E-02 & 4.92E-02 \\
 & SMA. & \textbf{9.18E-04} & \underline{9.20E-04} & 9.51E-04 & 9.64E-04 & 1.00E-03 & 1.16E-03 \\
 & FDE & \textbf{5.53E-02} & \underline{6.15E-02} & 6.20E-02 & 6.34E-02 & 6.77E-02 & 6.46E-02 \\
 & FD & \textbf{6.33E-02} & \underline{6.57E-02} & 6.63E-02 & 6.66E-02 & 6.99E-02 & 7.08E-02 \\
 & AED & \textbf{3.04E-02} & \underline{3.07E-02} & 3.19E-02 & 3.23E-02 & 3.37E-02 & 3.83E-02 \\ \bottomrule
\end{tabular}
\end{table}

\begin{table}[!htb]
\centering
\small
\setlength{\tabcolsep}{3pt}
\caption{Performance comparison under data split \revision{setting-2}. For each model, the best results are reported. The best-performing model is highlighted in bold, while the second-best is underlined. Note that this setting involves a risk of temporal data leakage due to overlapping trajectory segments; consequently, these results do not reflect the realistic generalization performance on unseen vessels.}
\label{tab:result}
\begin{tabular}{@{}cccccccc@{}}
\toprule
 &  & Mix\&Fix-Net & WPMixer & P\_sLSTM & PatchTST & Dlinear & FusionRNN\\ \midrule
\multirow{6}{*}{\rotatebox[origin=c]{90}{VVR}} & MSE & \textbf{3.86E-06} & 7.38E-04 & 5.33E-04 & 8.33E-04 & 1.16E-03 & \underline{9.43E-06} \\
 & MAE & \textbf{1.45E-03} & 1.59E-02 & 1.45E-02 & 1.72E-02 & 2.20E-02 & \underline{2.58E-03} \\
 & SMA. & \textbf{4.08E-03} & 5.04E-02 & 4.40E-02 & 5.47E-02 & 6.76E-02 & \underline{7.47E-03} \\
 & FDE & \textbf{1.64E-03} & 3.01E-02 & 2.68E-02 & 3.29E-02 & 4.15E-02 & \underline{3.13E-03} \\
 & FD & \textbf{3.72E-03} & 3.25E-02 & 2.85E-02 & 3.41E-02 & 4.23E-02 & \underline{5.35E-03} \\
 & AED & \textbf{1.16E-03} & 1.29E-02 & 1.19E-02 & 1.39E-02 & 1.75E-02 & \underline{2.06E-03} \\ \cmidrule(l){1-8}
\multirow{6}{*}{\rotatebox[origin=c]{90}{M3}} & MSE & \textbf{8.39E-05} & 2.82E-03 & 3.83E-03 & 2.90E-03 & 3.87E-03 & \underline{6.93E-04} \\
 & MAE & \textbf{7.24E-03} & 2.82E-02 & 4.03E-02 & 3.03E-02 & 3.46E-02 & \underline{1.75E-02} \\
 & SMA. & \textbf{4.13E-04} & \underline{1.46E-03} & 7.08E-03 & 1.58E-03 & 2.72E-03 & 3.64E-03 \\
 & FDE & \textbf{1.10E-02} & 5.98E-02 & 8.00E-02 & 6.46E-02 & 7.15E-02 & \underline{2.84E-02} \\
 & FD & \textbf{1.54E-02} & 6.05E-02 & 8.16E-02 & 6.52E-02 & 7.25E-02 & \underline{3.33E-02} \\
 & AED & \textbf{5.90E-03} & 2.25E-02 & 3.30E-02 & 2.41E-02 & 2.77E-02 & \underline{1.40E-02} \\ \cmidrule(l){1-8}
\multirow{6}{*}{\rotatebox[origin=c]{90}{TampaBay}} & MSE & \textbf{4.72E-04} & 2.58E-03 & 1.44E-03 & 2.59E-03 & 2.88E-03 & \underline{9.49E-04} \\
 & MAE & \textbf{1.76E-02} & 3.40E-02 & 2.66E-02 & 3.37E-02 & 3.92E-02 & \underline{2.53E-02} \\
 & SMA. & \textbf{4.14E-04} & 8.00E-04 & 6.23E-04 & 7.89E-04 & 9.12E-04 & \underline{6.06E-04} \\
 & FDE & \textbf{2.50E-02} & 5.59E-02 & 4.40E-02 & 5.57E-02 & 6.39E-02 & \underline{3.35E-02} \\
 & FD & \textbf{3.08E-02} & 5.89E-02 & 4.67E-02 & 5.84E-02 & 6.61E-02 & \underline{3.89E-02} \\
 & AED & \textbf{1.39E-02} & 2.70E-02 & 2.13E-02 & 2.68E-02 & 3.09E-02 & \underline{1.99E-02} \\ \bottomrule
\end{tabular}
\end{table}

\subsection{Results}
\label{results}
The quantitative performance of our Mix\&Fix-Net is compared with other baselines under both data-splitting settings. Table~\ref{tab:result_setting_2} and Table~\ref{tab:result}  show the results for setting-1 and setting-2, respectively. For the M3 dataset, performance is evaluated for each profile and then averaged across profiles. All baselines are reproduced using their publicly available code, with an extensive hyperparameter search using optuna, following their recommended ranges for the hyperparameters.\par

\begin{figure*}[!tb]
  \centering
  \subfloat{\includegraphics[width=0.95\textwidth]{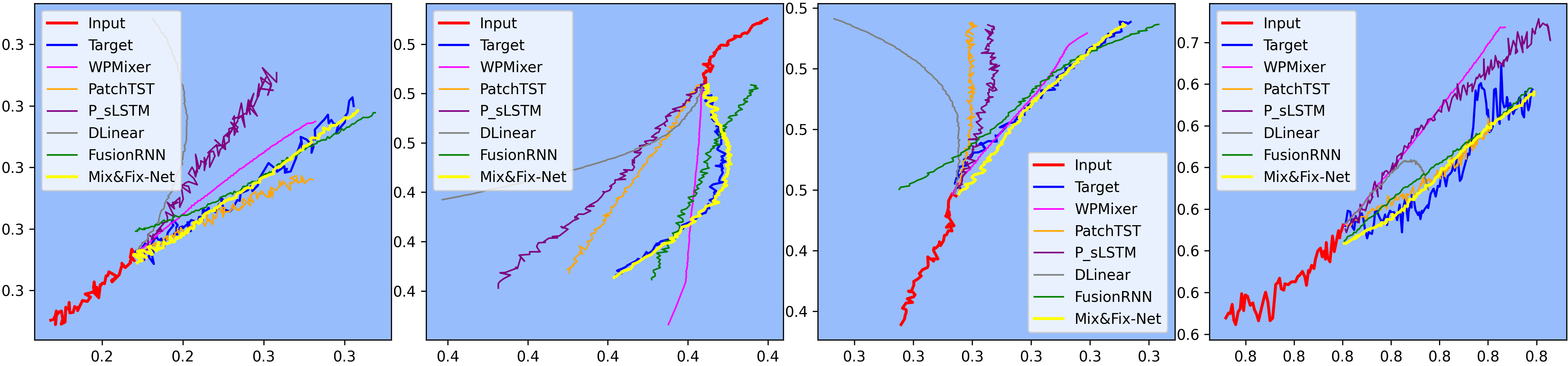}} \hfill
  \subfloat{\includegraphics[width=0.95\textwidth]{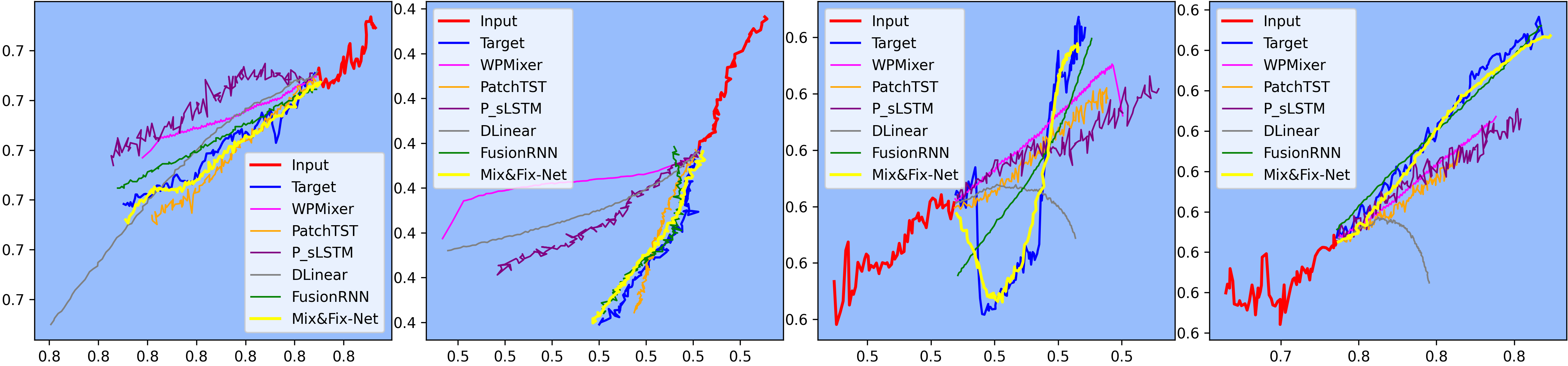}} \hfill
  \subfloat{\includegraphics[width=0.95\textwidth]{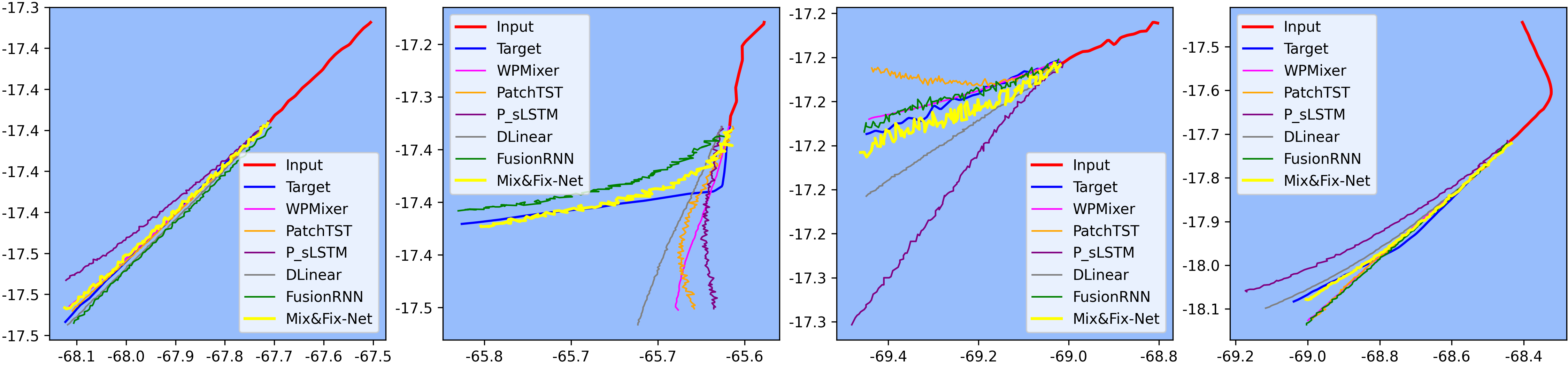}} \hfill
  \subfloat{\includegraphics[width=0.95\textwidth]{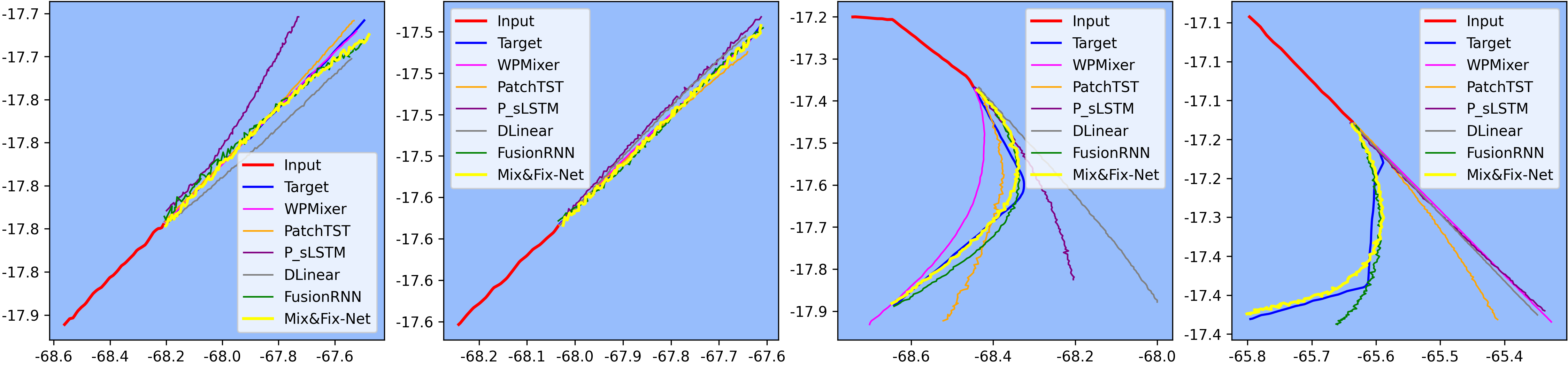}} 
  \caption{Visual comparison of the prediction among different models. Trajectories are randomly selected. The first two rows refer to the VVR dataset, while the last two rows refer to the M3 dataset. Predicted trajectories for the TampaBay dataset are shown in Figure~\ref{fig:route_prediction_tampabay_appendix} in the Appendix.}
  \label{fig:route_prediction}
\end{figure*}

\revision{Setting-1 (Table~\ref{tab:result_setting_2}) provides a rigorous and realistic evaluation of generalization, as the model is tested on entirely unseen vessels. As shown, Mix\&Fix-Net almost consistently outperforms the baselines, confirming that it maintains superior predictive power even when memorization is not possible. In contrast, Setting-2 (Table~\ref{tab:result}) uses random data splitting, under which the performance improvement of our model is considerably larger. However, this setting inherently carries a risk of data leakage due to overlapping trajectory segments between splits; its results should therefore be interpreted as a measure of the model's capacity to memorize and represent complex spatial patterns rather than to generalize to new vessels. For visualization, examples of randomly selected predictions are shown in Figure~\ref{fig:route_prediction} and Figure~\ref{fig:route_prediction_tampabay_appendix}. It is observed that the predicted trajectories of our model follow the ground-truth more accurately than the other baselines. However, the predicted trajectories exhibit some noise, which is analyzed in detail in~\ref{analysis_of_prediction_smoothness}.}

Among the baselines, the best-performing model after Mix\&Fix-Net is FusionRNN in terms of memorization capability (Setting-2). However, FusionRNN is outperformed by WPMixer and P\_sLSTM in terms of generalization capability (Setting-1). The VVR time-series dataset, generated by tracking vessels in videos, inherently contains noise, as illustrated in Figure~\ref{fig:route_prediction}. This noise limits the effectiveness of models that rely on decomposition, particularly WPMixer. In contrast, the M3 dataset is synthetic and noise-free. Therefore, WPMixer benefits from decomposition on this dataset compared to P\_sLSTM. Additionally, the simple MLP-based model DLinear performs the worst on almost all datasets and settings, despite its competitive or even superior performance compared to transformer-based models in standard time-series forecasting tasks.

\subsection{Computational Robustness and Efficiency}
\label{section_cost_robustness}
We evaluate all models with four different random seed values under the data-split setting-2, and report the average and the standard deviation values in Table \ref{tab:robustness}. The results indicate that our model is more robust compared to the other models. \par
Additionally, we present the computational complexity of our model in Table~\ref{tab:gflops}, measured by the average train iteration time, test iteration time, and GFLOPs (Giga Floating Point Operations) on the VVR dataset. While iteration time is a machine-dependent metric, GFLOPs is machine-independent. For training and testing, the batch size and number of iterations are identical for all models. Details of the machine specifications and hyperparameters are provided in Section~\ref{train_hyperparameters}.\par
The results in Table~\ref{tab:gflops} demonstrate that although our model incurs a significantly higher GFLOPs value than the competing methods, the train and test times remain comparable across all models. Several factors contribute to this increased GFLOPs value. Unlike WPMixer, P\_sLSTM, and PatchTST, our approach does not utilize the patching technique. Furthermore, we incorporate input padding at the beginning of the model, which triples the size of the data being processed. We also introduce the Residual Trajectory Adjuster module at the end of our model to refine coarse predictions. However, GPU parallelization enables efficient handling of these additional arithmetic operations. Additionally, as demonstrated in the ablation study (Section~\ref{ablation}), both the padding step and the Residual Trajectory Adjuster significantly enhance the accuracy of trajectory prediction, thereby justifying the additional computational overhead.\par
Further, our method is suitable for real-time deployment, as demonstrated in this video \footnote{\url{https://youtu.be/CgXx3Pa-xzY}}. \revision{To assess practical deployability, we measure inference latency on a single NVIDIA RTX 4090: each test batch (size 128) takes 0.030~s, i.e., about 0.23~ms per trajectory prediction (over 4{,}000 predictions/s), each spanning a 120-step horizon. The high GFLOPs thus does not translate into a proportional latency cost. This throughput is well within the shore-station and port-authority scenarios we target, which use GPU-class hardware. For more constrained on-board or edge settings, a lightweight variant with a smaller embedding dimension $d$ (Figure~\ref{fig:d_model_sensitivity}) reduces GFLOPs substantially at a modest accuracy cost. The accuracy–efficiency trade-off therefore remains favorable, as the practical cost is the wall-clock latency, which stays within real-time.}

\begin{table}[]
\centering
\footnotesize
\caption{Robustness comparison. Each model is evaluated using four random seed values.}
\label{tab:robustness}
\vspace{1mm}
\begin{tabular}{@{}llcc|cc|cc@{}}
\toprule
 &  & \multicolumn{2}{c}{Mix\&Fix-Net} & \multicolumn{2}{c}{WPMixer} & \multicolumn{2}{c}{P\_sLSTM} \\
 &  & Mean & Std. & Mean & Std. & Mean & Std. \\
\midrule
\multirow{6}{*}{VVR}
 & MSE   & 4.12E-06 & 2.09E-07 & 7.65E-04 & 1.89E-05 & 5.33E-04 & 4.51E-06 \\
 & MAE   & 1.46E-03 & 3.29E-05 & 1.61E-02 & 1.69E-04 & 1.43E-02 & 1.46E-04 \\
 & SMAPE & 4.11E-03 & 9.49E-05 & 5.12E-02 & 5.68E-04 & 4.37E-02 & 2.59E-04 \\
 & FDE   & 1.64E-03 & 2.39E-05 & 3.09E-02 & 5.94E-04 & 2.67E-02 & 9.26E-05 \\
 & FD    & 3.71E-03 & 4.89E-05 & 3.35E-02 & 7.89E-04 & 2.84E-02 & 1.05E-04 \\
 & AED   & 1.17E-03 & 2.62E-05 & 1.31E-02 & 1.33E-04 & 1.17E-02 & 1.02E-04 \\
\midrule
\multirow{6}{*}{M3}
 & MSE   & 1.60E-04 & 1.14E-04 & 2.94E-03 & 8.35E-05 & 3.67E-03 & 2.97E-04 \\
 & MAE   & 9.14E-03 & 2.75E-03 & 2.97E-02 & 1.41E-03 & 3.92E-02 & 1.08E-03 \\
 & SMAPE & 4.85E-04 & 1.21E-04 & 1.45E-03 & 3.92E-05 & 6.83E-03 & 6.98E-04 \\
 & FDE   & 1.46E-02 & 5.21E-03 & 6.11E-02 & 1.29E-03 & 7.75E-02 & 1.94E-03 \\
 & FD    & 1.98E-02 & 5.86E-03 & 6.18E-02 & 1.31E-03 & 7.94E-02 & 1.79E-03 \\
 & AED   & 7.51E-03 & 2.29E-03 & 2.38E-02 & 1.06E-03 & 3.21E-02 & 9.81E-04 \\
 \midrule
\multirow{6}{*}{TampaB.}
 & MSE & 4.61E-04 & 2.57E-05 & 2.60E-03 & 2.08E-05 & 1.50E-03 & 6.15E-05 \\
 & MAE & 1.79E-02 & 4.02E-04 & 3.41E-02 & 1.38E-04 & 2.71E-02 & 5.12E-04 \\
 & SMAPE & 4.21E-04 & 1.00E-05 & 8.03E-04 & 3.77E-06 & 6.34E-04 & 1.26E-05 \\
 & FDE & 2.43E-02 & 7.81E-04 & 5.59E-02 & 1.36E-04 & 4.49E-02 & 9.50E-04 \\
 & FD & 3.04E-02 & 7.92E-04 & 5.90E-02 & 1.47E-04 & 4.75E-02 & 9.07E-04 \\
 & AED & 1.40E-02 & 3.22E-04 & 2.71E-02 & 1.06E-04 & 2.17E-02 & 4.06E-04 \\
 \bottomrule
\end{tabular}
\vspace{1mm}
\vspace{1mm}
\begin{tabular}{@{}llcc|cc|cc@{}}
\toprule
 &  & \multicolumn{2}{c}{PatchTST} & \multicolumn{2}{c}{DLinear} & \multicolumn{2}{c}{FusionRNN}\\
 &  & Mean & Std. & Mean & Std. & Mean & Std. \\
\midrule
\multirow{6}{*}{VVR}
 & MSE   & 8.32E-04 & 5.34E-06 & 1.16E-03 & 1.65E-07 & 1.02E-05 & 9.76E-07 \\
 & MAE   & 1.72E-02 & 5.52E-05 & 2.20E-02 & 6.66E-05 & 2.58E-03 & 1.19E-04 \\
 & SMAPE & 5.46E-02 & 1.96E-04 & 6.77E-02 & 2.64E-04 & 7.49E-03 & 3.41E-04 \\
 & FDE   & 3.29E-02 & 1.03E-04 & 4.15E-02 & 1.23E-05 & 3.14E-03 & 1.55E-04 \\
 & FD    & 3.41E-02 & 1.45E-04 & 4.23E-02 & 1.23E-05 & 5.39E-03 & 1.76E-04 \\
 & AED   & 1.39E-02 & 4.93E-05 & 1.75E-02 & 4.78E-05 & 2.06E-03 & 9.65E-05 \\
\midrule
\multirow{6}{*}{M3}
 & MSE   & 3.00E-03 & 6.52E-05 & 3.95E-03 & 1.30E-04 & 1.90E-03 & 1.38E-03 \\
 & MAE   & 3.09E-02 & 6.84E-04 & 3.53E-02 & 1.25E-03 & 2.68E-02 & 1.10E-02 \\
 & SMAPE & 1.58E-03 & 8.96E-05 & 3.20E-03 & 8.63E-04 & 9.18E-03 & 5.80E-03 \\
 & FDE   & 6.55E-02 & 1.31E-03 & 7.28E-02 & 1.94E-03 & 3.63E-02 & 9.99E-03 \\
 & FD    & 6.62E-02 & 1.30E-03 & 7.48E-02 & 4.08E-03 & 4.34E-02 & 1.20E-02 \\
 & AED   & 2.47E-02 & 5.57E-04 & 2.83E-02 & 1.04E-03 & 2.15E-02 & 8.82E-03 \\
\midrule
\multirow{6}{*}{TampaB.}
 & MSE & 2.97E-03 & 1.75E-04 & 2.88E-03 & 2.09E-06 & 8.72E-04 & 5.39E-05 \\
 & MAE & 3.69E-02 & 1.85E-03 & 3.93E-02 & 2.83E-04 & 2.45E-02 & 5.61E-04 \\
 & SMAPE & 8.67E-04 & 4.47E-05 & 9.12E-04 & 5.69E-06 & 5.86E-04 & 1.36E-05 \\
 & FDE & 5.89E-02 & 9.98E-04 & 6.41E-02 & 4.39E-04 & 3.22E-02 & 8.79E-04 \\
 & FD & 6.20E-02 & 1.16E-03 & 6.64E-02 & 4.24E-04 & 3.77E-02 & 8.48E-04 \\
 & AED & 2.93E-02 & 1.50E-03 & 3.10E-02 & 2.11E-04 & 1.92E-02 & 4.52E-04 \\
 \bottomrule
 
\end{tabular}
\end{table}

\begin{table}[!htb]
\centering
\small
\caption{Computational complexity measured by average iteration time (in seconds) for NVIDIA 4090 GPU and GFLOPs for the VVR dataset.}
\label{tab:gflops}
\begin{tabular}{@{}lccc@{}}
\toprule
 & Train Time & Test time & GFLOPs \\ \midrule
Mix\&Fix-Net & 0.086 & 0.030 & 1136 \\
WPMixer & 0.011 & 0.005 & 11 \\
P\_sLSTM & 0.029 & 0.010 & 3 \\
PatchTST & 0.010 & 0.006 & 35 \\
Dlinear & 0.008 & 0.004 & 0.03 \\
FusionRNN & 0.008 & 0.003 & 14 \\ \bottomrule
\end{tabular}
\end{table}

\begin{table*}[!htb]
\centering
\footnotesize
\caption{Relative change in six metrics with respect to the base case (Case-1). All values are reported as percentages, \revlast{while the corresponding absolute values are shown in Table~\ref{tab:ablation_abs_value}}. Symbol $\spadesuit$ in the Head column refers to our proposed head layer design, while symbol $\diamondsuit$ refers to the regular head layer design for time series forecasting models. The "$D$" in the RevIN column refers only to denormalization, whereas "$ND$" refers to both normalization and denormalization. RTA refers to the Residual Trajectory Adjuster module.}
\label{tab:ablation}
\begin{tabular}{@{}ccccccc|cccccc@{}}
\toprule
\rotatebox{90}{Case} & \rotatebox{90}{RTA} & \rotatebox{90}{Emb. Mx} & \rotatebox{90}{Tm. Mx} & \rotatebox{90}{Head} & \rotatebox{90}{Padding} & \rotatebox{90}{RevIN} & \rotatebox{90}{$\Delta$MSE} & \rotatebox{90}{$\Delta$MAE} & \rotatebox{90}{$\Delta$SMAPE} & \rotatebox{90}{$\Delta$FDE} & \rotatebox{90}{$\Delta$FD} & \rotatebox{90}{$\Delta$AED} \\ \midrule
1 & \checkmark & \checkmark & \checkmark & $\spadesuit$ & \checkmark & $D$ & 0 & 0 & 0 & 0 & 0 & 0 \\
2 & $\times$ & \checkmark & \checkmark & $\spadesuit$ & \checkmark & $D$ & 14 & 5 & 4 & 7 & 2 & 5 \\
3 & \checkmark & \checkmark & $\times$ & $\spadesuit$ & \checkmark & $D$ & 157906 & 4562 & 3932 & 5312 & 2365 & 4565 \\
4 & \checkmark & $\times$ & \checkmark & $\spadesuit$ & \checkmark & $D$ & 322 & 131 & 140 & 163 & 87 & 131 \\
5 & \checkmark & \checkmark & \checkmark & $\spadesuit$ & \checkmark & $ND$ & 231 & 67 & 66 & 98 & 44 & 68 \\
6 & \checkmark & \checkmark & \checkmark & $\spadesuit$ & \checkmark & - & 39 & 23 & 26 & 27 & 14 & 22 \\
7 & \checkmark & \checkmark & \checkmark & $\spadesuit$ & $\times$ & $D$ & 208 & 99 & 98 & 128 & 63 & 101 \\
8 & \checkmark & \checkmark & \checkmark & $\diamondsuit$ & \checkmark & $D$ & 495 & 180 & 193 & 235 & 114 & 182 \\ \bottomrule
\end{tabular}
\end{table*}

\section{Ablation Study}
\label{ablation}
In this section, we present ablation studies on the contributions of individual modules. Additional ablation studies on the role of non-positional features and the sensitivity analysis of the hyperparameters are demonstrated in~\ref{additional_ablation_studies_appendix}. For all experiments, data-splitting setting-2 is employed with independent hyperparameter tuning.

\subsection{Impact of different modules in Mix\textnormal{\&}Fix-Net}
\label{sec:modular_ablation}
In this section, we evaluate the impact of different modules on the overall performance of our model. We conduct experiments with eight cases, where in each case, we remove or disable a different module. For fairness, the hyperparameters are tuned separately for every case. The results, presented in Table \ref{tab:ablation}, report the percentage change of six metrics relative to the base case (Case-1). The base case (Case-1) corresponds to our proposed Mix\&Fix-Net model.

In Case-2, the Residual Trajectory Adjuster is removed. Removing it increases the error across all metrics, with MSE, MAE, SMAPE, FDE, FD, and AED rising by about 2–14 percent. In Case-4, the embedding mixer is removed, which severely degrades performance, with MSE increasing by 322 percent.

To mitigate distributional shift in time series, RevIN (Reversible Instance Normalization) was introduced in \cite{revin}. It applies instance normalization to the input at the beginning of the model and instance denormalization to the output at the end. However, in our model, we employ only the instance denormalization after the Base Trajectory Predictor module. This design choice reflects the fact that vessel trajectory prediction depends not only on the trajectory pattern but also on the vessel’s absolute position in the sea area. Instance normalization eliminates this localized positional information. Therefore, we omit instance normalization but retain instance denormalization. The reason for retaining the instance denormalization is to restore the scale and distribution of the predicted trajectory. This ensures that the output remains consistent with the original spatial domain. This effect is validated in Case-5 and Case-6. In Case-5, when both normalization and denormalization are applied, performance degrades significantly. In Case-6, when both are removed, performance also declines, though less severely. These results show that denormalization alone is beneficial while normalization is detrimental for this task.

In Case-7, we remove the padding step. The results show a clear drop in performance, confirming the importance of padding in our architecture.

Finally, the head layer of Mix\&Fix-Net differs from that of other time series forecasting models such as WPMixer, P\_sLSTM, and PatchTST, which all share a common head design. To evaluate this difference, we replace our head with WPMixer’s head in Case-8. The results also show performance degradation. In WPMixer, the head transformation is defined as $\mathbb{R}^{\cdots \times (P \cdot d)} \xrightarrow{Linear} \mathbb{R}^{\cdots \times T}$, where $P$ and $T$ are the number of patches and predicted sequences, and $d$ is the embedding dimension. This transformation is applied per channel, making it suitable for standard time series prediction where each channel is predicted separately. However, in the trajectory prediction task, the prediction of $x$-coordinate channel is not performed solely, because the value of the $x$-coordinate depends on the value of the $y$-coordinate. The details of our head layer are described in Section~\ref{head}.

From Table \ref{tab:ablation}, it is evident that the most critical component of our model is the Temporal Mixer module, as shown in Case-3. The Temporal Mixer captures temporal dependencies among data points. Without it, the model cannot learn the trajectory pattern, leading to a dramatic loss in performance.

\section{Difference with the Existing Models} Although vessel trajectory prediction can be formulated as a time series forecasting problem, our proposed Mix\&Fix-Net departs from conventional forecasting architectures in several key ways. These differences are necessary to handle the trajectory prediction task, as validated in our ablation study Section~\ref{sec:modular_ablation}.

In the traditional Head layer $(\Diamond)$, all future data points are projected channel-wise. On the contrary, our Head layer $(\spadesuit)$ projects both positional channels for each future data point at the same time. 

Additionally, padding the input has not been explored before for the time series forecasting task. In the ablation study, we show the effectiveness of padding in the trajectory prediction.

Employing both reversible instance normalization and denormalization is a well-established approach in time series forecasting. However, in our model, we only employ denormalization. In Table~\ref{tab:ablation}, we demonstrate that employing RevIN-normalization degrades the trajectory prediction performance.

Moreover, we employ a dual-stage mixer architecture, where the initial mixer predicts the coarse prediction, and the following mixer refines the earlier coarse prediction. This approach is also novel and effective in terms of the refinement stage.

\section{Limitations and Observations}
\label{limitations}
Despite the predictive accuracy of Mix\&Fix-Net, certain limitations regarding video-based data must be acknowledged. Our VVR dataset was recorded from the broadcast webcam stream on the Internet, resulting in relatively low-resolution footage. Additionally, the dataset's temporal coverage is limited. Moreover, the trajectories extracted from videos exhibit less smoothness than AIS data due to pixel-level tracking noise and perspective distortion inherent in fixed-camera observations. Consequently, these predicted trajectories may not always strictly conform to idealized physical vessel kinematics. Future research will explore integrating temporal smoothing techniques, \revlast{such as smoothness regularization}, with homography-based coordinate mapping to better align image-plane observations with realistic geographical motion patterns. \par

The findings in our work have significant practical implications for governmental authorities worldwide. National coast guards and navies can benefit from the proposed framework through enhanced surveillance and monitoring capabilities for small vessels operating without AIS. Furthermore, vision-based trajectory prediction can support environmental monitoring activities conducted by environmental and oceanographic institutions. The model can also contribute to traffic optimization and digital maritime infrastructure initiatives led by maritime administrations through improved predictive analytics for vessel movement patterns. \par

\section{Conclusion}
Vessel trajectory prediction is a complex task as the trajectory of the vessel is influenced by various factors such as ocean currents, weather conditions, and navigational constraints at sea. The widespread availability of AIS data has enabled substantial research in the trajectory prediction domain. However, not all vessels are equipped with AIS, and obstacles along the trajectory, such as restricted zones or dynamic hazards, cannot be identified solely through AIS data. 
To address these challenges, we propose Mix\&Fix-Net, a novel dual-stage MLP-mixer-based model that is applicable to both AIS and non-AIS data. Additionally, we introduce VVR, a video-based dataset developed specifically for vessel trajectory prediction. Our model is evaluated on both the non-AIS dataset (VVR) and the AIS datasets (M3, TampaBay) using six metrics: MSE, MAE, SMAPE, FDE, FD, and AED, under two different data-splitting settings. Experimental results demonstrate that Mix\&Fix-Net consistently outperforms baseline methods \revision{across most metrics and datasets}. Through a detailed ablation study, we further examine the contribution of individual modules within our model and highlight its advantages over conventional time series forecasting-based models. These findings underscore the potential of Mix\&Fix-Net as a robust and \revlast{effective} framework for advancing vessel trajectory prediction.


\section*{Acknowledgments}
The authors would like to thank Awwab Ahmed, Nguyen Hong, and Emily Wyatt for their efforts in annotating vessels for the VVR dataset. 

This work was supported by the U.S. National Institute of Food and Agriculture under Grant 2023-67019-38829. Additionally, Bora San Turgut was supported by the Scientific and Technological Research Council of Türkiye (TUBITAK) during this work.


\bibliographystyle{elsarticle-num} 
\bibliography{references}






\appendix
\pagebreak

\section{Vessel Detection and Tracking Implementation Details}
\label{vessel_detection_and_tracking_imp_details}
Our Video-based Vessel Route (VVR) dataset consists of six video recordings, detailed in Table~\ref{tab:video_dataset_details}. The first two videos are used to train and evaluate the vessel detection model, while the remaining four videos are used to generate the vessel's time-series trajectories. Although the pretrained YOLOv11 model (YOLOv11x variant with pretrained weights) can detect ships or vessels, we observe frequent missed detections for small vessels in our recordings. Therefore, we fine-tune the pretrained model on our domain-specific data.\par

\textbf{YOLO Fine-Tuning:}
The first two videos are converted into image frames. All vessels appearing in these frames were manually annotated with bounding boxes. The total number of image frames is split into a train and a test set with a 7:3 ratio. YOLO is then fine-tuned and evaluated on the train and test sets, respectively. To fine-tune it, we employ an initial learning rate of 0.0002, a batch size of 64, and 100 epochs. Additionally, we employ basic augmentation (e.g., scaling, flipping, color shifting) during training; augmentation parameters are shown in Table~\ref{tab:yolo_augmentation_params}.\par

\textbf{Multi-Object Tracking:}
To track vessels, we use the ByteTrack algorithm with the fine-tuned YOLO model. The experimental settings of the tracker are shown in Table~\ref{tab:bytetracker_params}. The tracker assigns a unique vessel ID to each detected object across frames. The center coordinates of the bounding boxes are extracted to generate the vessel trajectory time-series data.

\begin{table}[!htb]
\centering
\footnotesize
\setlength{\tabcolsep}{8pt}
\caption{Data augmentation parameters used for YOLOv11 fine-tuning on the VVR dataset.}
\label{tab:yolo_augmentation_params}
\begin{tabular}{@{}llc@{}}
\toprule
Parameter & Description & Value \\ \midrule
\texttt{hsv\_h}     & Image HSV-Hue augmentation (fraction)       & 0.015 \\
\texttt{hsv\_s}     & Image HSV-Saturation augmentation (fraction) & 0.7   \\
\texttt{hsv\_v}     & Image HSV-Value augmentation (fraction)      & 0.4   \\
\texttt{translate} & Image translation (+/- fraction)            & 0.1   \\
\texttt{scale}     & Image scale (+/- gain)                      & 0.5   \\
\texttt{fliplr}    & Image flip left-right (probability)         & 0.5   \\
\texttt{degrees}   & Image rotation (+/- deg)                    & 0.0   \\
\texttt{shear}     & Image shear (+/- deg)                       & 0.0   \\
\texttt{flipud}    & Image flip up-down (probability)            & 0.0   \\ \bottomrule
\end{tabular}
\end{table}

\begin{table}[!htb]
\centering
\footnotesize
\setlength{\tabcolsep}{8pt}
\caption{Experimental settings of the ByteTrack algorithm. Parameter definitions follow the standard implementation in \cite{ultralytics_tracking_2026}.}
\label{tab:bytetracker_params}
\begin{tabular}{@{}llc@{}}
\toprule
    Parameter & Description & Value \\ \midrule
    \texttt{track\_high\_thresh} & Threshold for the first association (high score)  & 0.5   \\
    \texttt{track\_low\_thresh}  & Threshold for the second association (low score)   & 0.1   \\
    \texttt{new\_track\_thresh}   & Threshold for initiating a new track              & 0.5   \\
    \texttt{track\_buffer}       & Number of frames to retain a lost track           & 600   \\
    \texttt{match\_thresh}       & Intersection over Union (IoU) matching threshold  & 0.95  \\ \bottomrule
\end{tabular}
\end{table}

\begin{table}[!htb]
\centering
\footnotesize
\setlength{\tabcolsep}{4pt}
\caption{Hyperparameter settings of the proposed model across benchmark datasets under both experimental settings. Here, $L$ and $T$ denote the lookback window length and the prediction window length, respectively. Additionally, $d$, $t_f$, $d_f$, drop. refer to the embedding dimension, temporal expansion factor, embedding expansion factor, and dropout value, respectively.}
\label{tab:hyperparams_tables}
\begin{tabular}{@{}c|c|cccccccccc@{}}
\toprule
\revision{Setting} & Data & $L$ & $T$ & Initial lr. & Batch & $d$ & $t_f$ & $d_f$ & drop. & Epochs & Patience \\ \midrule
\multirow{3}{*}{1} & VVR & 60 & 120 & 8.13E-05 & 128 & 16 & 5 & 2 & 0.05 & 50 & 10 \\
 & M3 & 60 & 120 & 0.004887 & 512 & 16 & 5 & 1 & 0.05 & 50 & 10 \\
 & TampaBay & 60 & 120 & 6.72E-05 & 128 & 16 & 1 & 3 & 0.05 & 50 & 10 \\ \midrule
\multirow{3}{*}{2} & VVR & 60 & 120 & 0.001172 & 128 & 256 & 5 & 5 & 0 & 50 & 10 \\
 & M3 & 60 & 120 & 0.001392 & 512 & 256 & 3 & 5 & 0 & 50 & 10 \\
 & TampaBay & 60 & 120 & 8.18E-05 & 128 & 128 & 5 & 2 & 0.1 & 50 & 10 \\ \bottomrule

\end{tabular}
\end{table}

\begin{figure*}[!htb]
  \centering
  \subfloat{\includegraphics[width=0.95\textwidth]{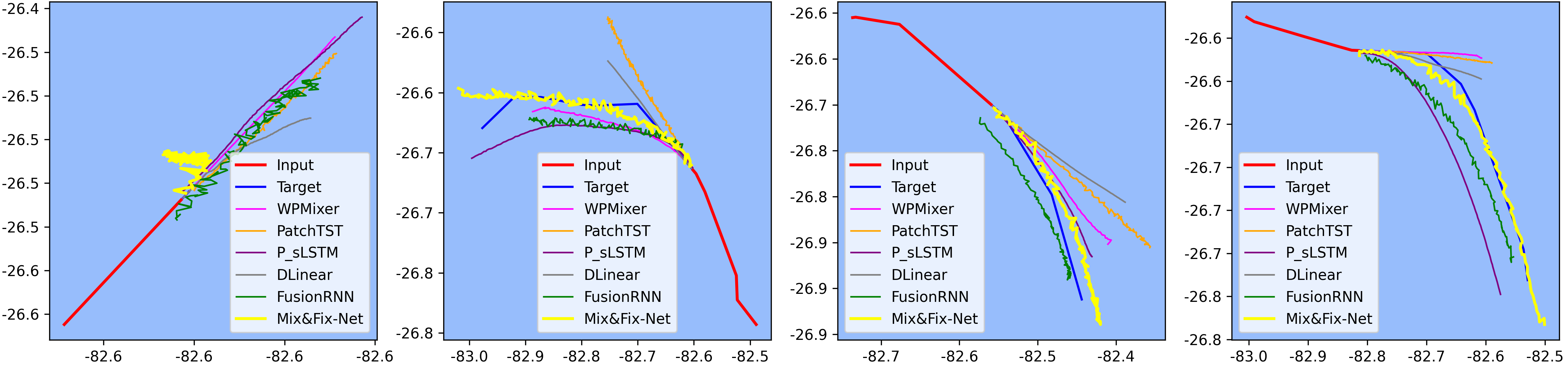}} \hfill
  \subfloat{\includegraphics[width=0.95\textwidth]{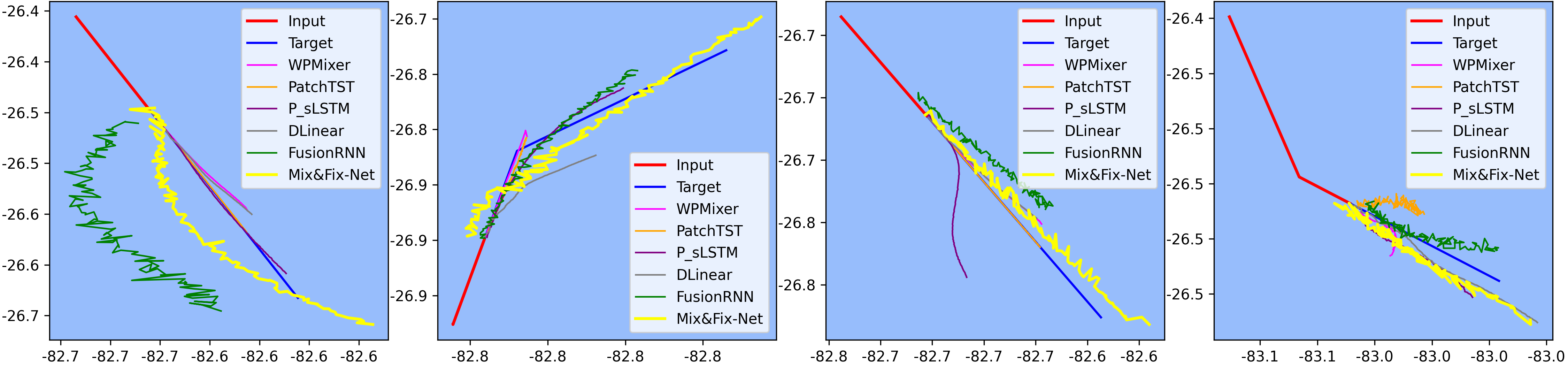}}  
  \caption{Visual comparison of the predicted trajectory among different models for the TampaBay dataset. Trajectories are randomly selected}
  \label{fig:route_prediction_tampabay_appendix}
\end{figure*}

\section{Analysis of Prediction Smoothness}
\label{analysis_of_prediction_smoothness}
As observed in Figure~\ref{fig:route_prediction} and Figure~\ref{fig:route_prediction_tampabay_appendix}, the trajectories predicted by Mix\&Fix-Net exhibit slightly more local irregularity than those of some baselines. This behavior arises from two distinct factors, one data-related and one architectural:

First, the VVR dataset is inherently noisy due to abrupt bounding-box variations from the detector and tracker, and since the training data itself is noisy, this noise propagates into the predictions of all models. \par

Second, even on the cleaner TampaBay and M3 datasets, our predictions show slight irregularity, and this is due to architectural design. The head layer of our model predicts the position at each future timestep individually from a dedicated hidden representation:
\begin{align}
    \R{1 \times (C_i \cdot d)} \xrightarrow{\text{Linear}} \R{1 \times C_i}.
\end{align}
\noindent Here, $\R{1 \times (C_i \cdot d)}$ represents the hidden feature vectors for time step $t$, and $\R{1 \times C_i}$ represents the corresponding output transformation at time step $t$. From the $C_i$ features, we only consider the last two features ($C_o = 2$) as our final positional coordinates (x/y or lat/lon.). Conversely, the baselines derive all $T$ predicted positions from a single shared hidden representation. For example, the head layer of WPMixer can be represented as,
\begin{align}
    \mathbb{R}^{\cdots \times (P \cdot d)} \xrightarrow{\text{Linear}} \mathbb{R}^{\cdots \times T}
\end{align}
where $P$ and $d$ refer to the patch number and embedding dimension, respectively. Here, the same hidden representation $\mathbb{R}^{\cdots \times (P \cdot d)}$ is shared across all $T$ predicted time steps of a given channel, and a single linear projection produces the full output sequence at once. Since this shared representation implicitly couples adjacent outputs, this yields globally smooth curves. But our per-timestep design does not impose this coupling.\par

The benefit of our design, however, is significant: because each predicted point has its own hidden representation, our predicted trajectories follow the ground-truth curvature (especially along sharp turns) far more accurately than the baselines, as visible in Figure~\ref{fig:route_prediction} and Figure~\ref{fig:route_prediction_tampabay_appendix}. This benefit is also confirmed quantitatively in Table~\ref{tab:ablation} (Case-1 vs. Case-8), where replacing our head with the conventional head (time-series model) increases MSE, MAE, SMAPE, FDE, FD, and AED by 495\%, 180\%, 193\%, 235\%, 114\%, and 182\%, respectively. Thus, although the predicted trajectory appears slightly noisier, it tracks the true trajectory much more accurately, which is the primary objective of trajectory prediction.

\section{Additional Ablation Studies}
\label{additional_ablation_studies_appendix}
This section presents the impact on non-positional features and hyperparameter sensitivity analysis. For all experiments, dataset split setting-2 is employed, and the hyperparameter is tuned using Optuna.

\subsection{Impact of non-positional features}
\label{ablation_non_positional_features}
The VVR dataset comprises nine features: $M_{sin}$, $M_{cos}$, $H_{sin}$, $H_{cos}$, $D_{sin}$, $D_{cos}$, $D_{disp}$, $x$-coordinate, and $y$-coordinate. In contrast, the M3 dataset includes an additional feature, Profile-ID. To assess the influence of non-positional features, we evaluated the model using three configurations: positional features only ($C_i=2$), the feature set excluding day-of-week encodings ($C_i=7$), and the full feature set ($C_i=9$). The results, presented in Table~\ref{tab:num_channels}, show that while positional data provides a baseline, the inclusion of auxiliary features significantly reduces error. Notably, the transition from $C_i=7$ to $C_i=9$ demonstrates that even with limited temporal coverage, the day-of-week features provide beneficial high-dimensional context for the model.
This benefit does not rely only on capturing the periodic behavioral patterns; rather, these sinusoidal encodings function as expressive temporal positional encodings, providing a unique and structured temporal stamp for each data point.

\begin{table}[!htb]
\centering
\small
\caption{Impact of the non-positional features. $C_i$ denotes the number of input features: $C_i = 2$ (positional only), $C_i = 7$ (all features excluding $D_{sin}/D_{cos}$), 
and $C_i = 9$ (full feature set).}
\label{tab:num_channels}
\begin{tabular}{@{}llccc@{}}
\toprule
 & $C_i$  & 2 & 7 & 9 \\ \midrule
\multirow{6}{*}{VVR} 
 & MSE & 2.16E-05 & 4.63E-06 & \textbf{3.86E-06} \\ 
 & MAE & 3.50E-03 & 1.49E-03 & \textbf{1.45E-03} \\
 & SMAPE & 1.02E-02 & 4.18E-03 & \textbf{4.08E-03} \\
 & FDE & 4.92E-03 & 1.70E-03 & \textbf{1.64E-03} \\
 & FD & 7.08E-03 & 3.79E-03 & \textbf{3.72E-03} \\
 & AED & 2.82E-03 & 1.19E-03 & \textbf{1.16E-03} \\ \bottomrule
\end{tabular}
\end{table}

\subsection{Sensitivity Analysis and Temporal Scale Justification}
\label{hyper_sensiti_appendix}
\begin{figure}[!htb]
    \centering
    \includegraphics[width=0.6\columnwidth]{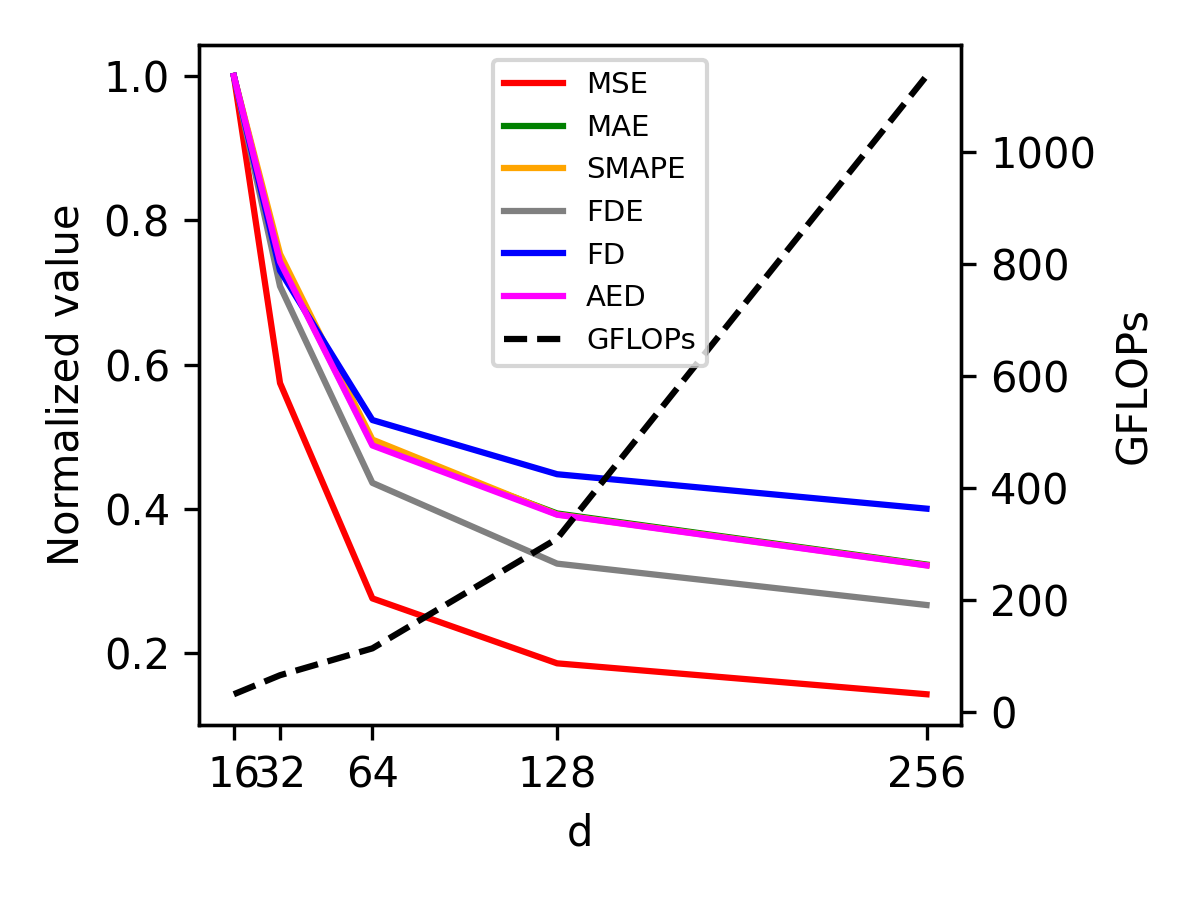}
    \caption{Impact of embedding dimension $d$ on prediction performance and computational cost for the VVR dataset.}
    \label{fig:d_model_sensitivity}
\end{figure}

In our model, one of the critical hyperparameters is the embedding dimension $d$. To assess its sensitivity, we evaluate our model using $d \in \{16, 32, 64, 128, 256\}$. As shown in Figure~\ref{fig:d_model_sensitivity}, performance metrics generally improve as $d$ increases. This occurs because the model captures more fine-grained information with higher embedding dimensions. While performance improves gradually with larger $d$, this comes at the cost of significantly higher GFLOPs. Considering this trade-off, we restrict $d$ to values not exceeding 256.\par

\begin{figure}[!htb]
    \centering
    \includegraphics[width=0.6\columnwidth]{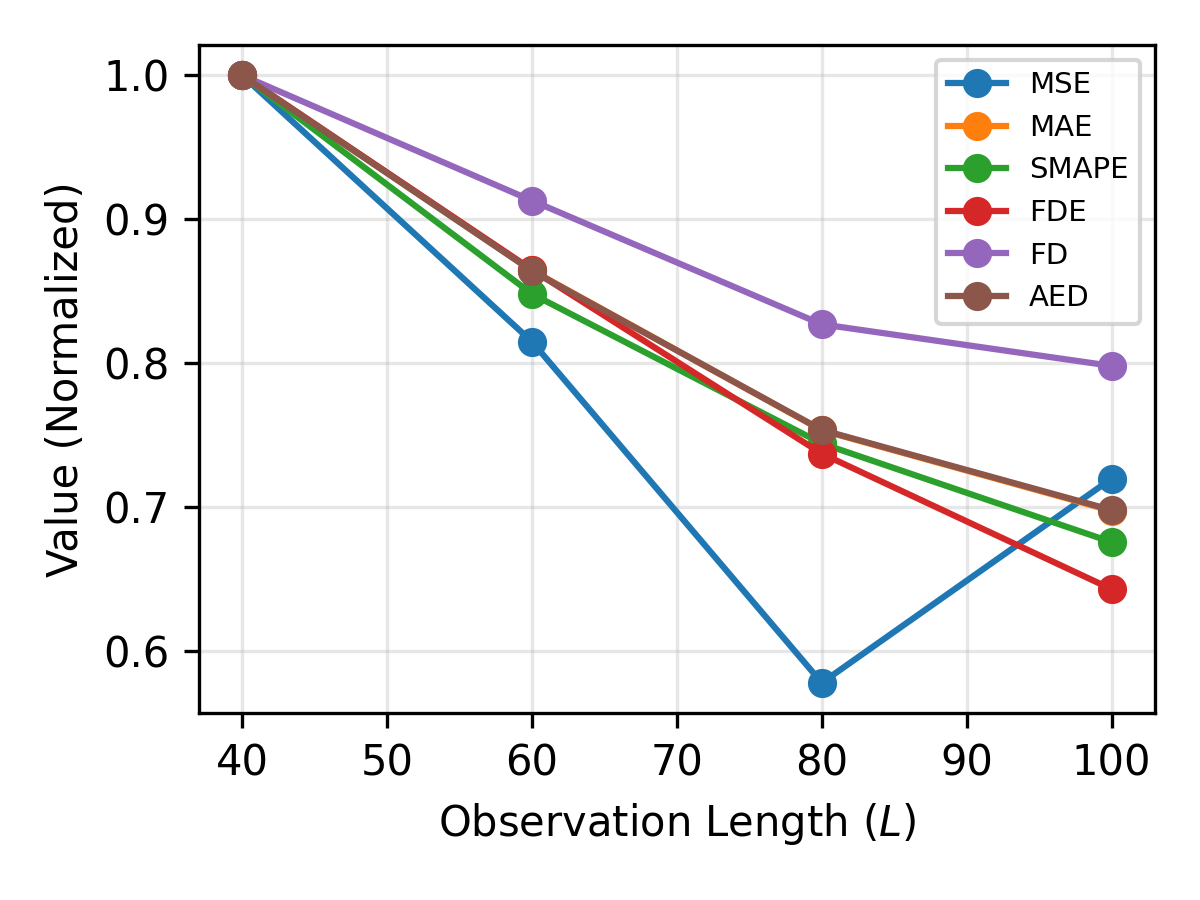}
    \caption{Impact of observation window $L$ on prediction performance for the VVR dataset.}
    \label{fig:L_sensitivity}
\end{figure}

\begin{figure}[!htb]
  \centering
  \subfloat[VVR]{%
    \includegraphics[width=0.49\textwidth]{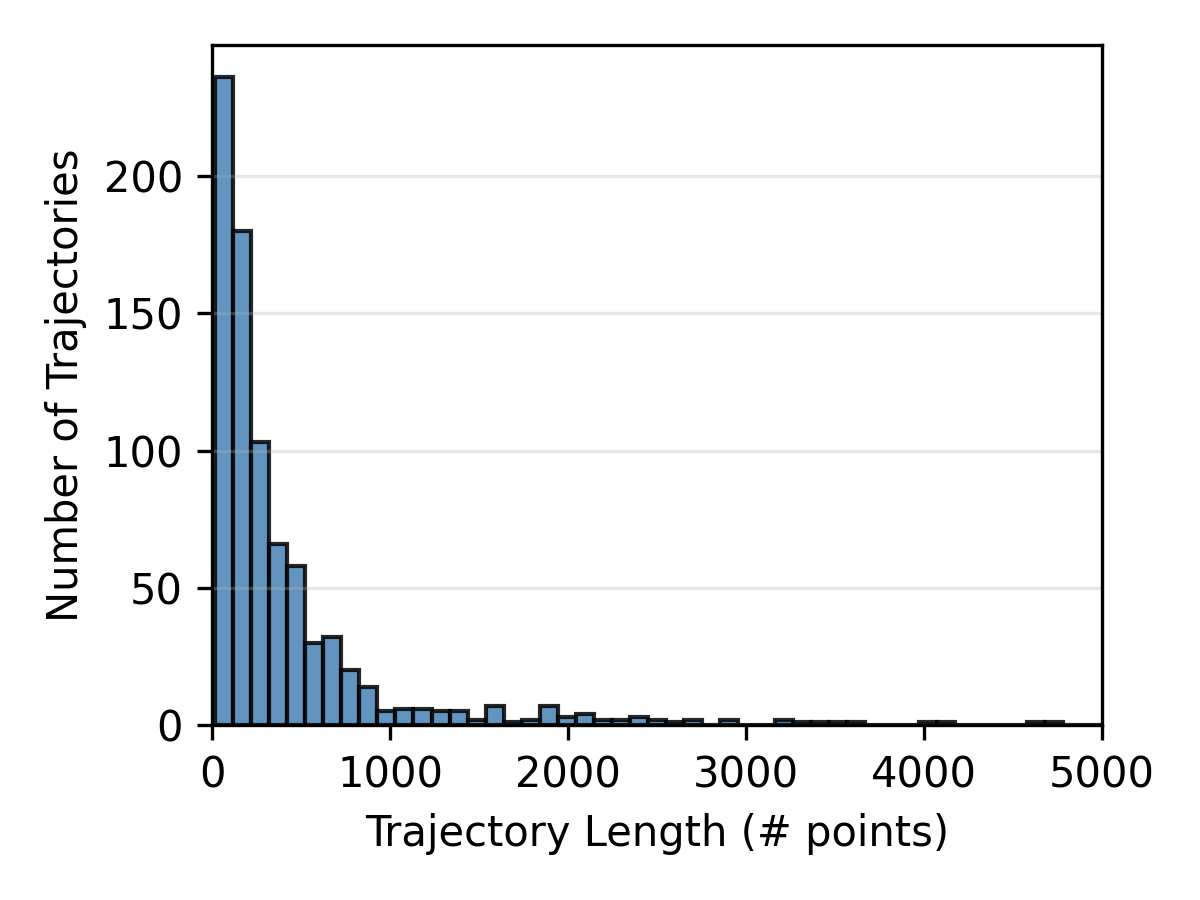}%
    \label{fig:vvr_hist}}
  \hfill
  \subfloat[TampaBay]{%
    \includegraphics[width=0.49\textwidth]{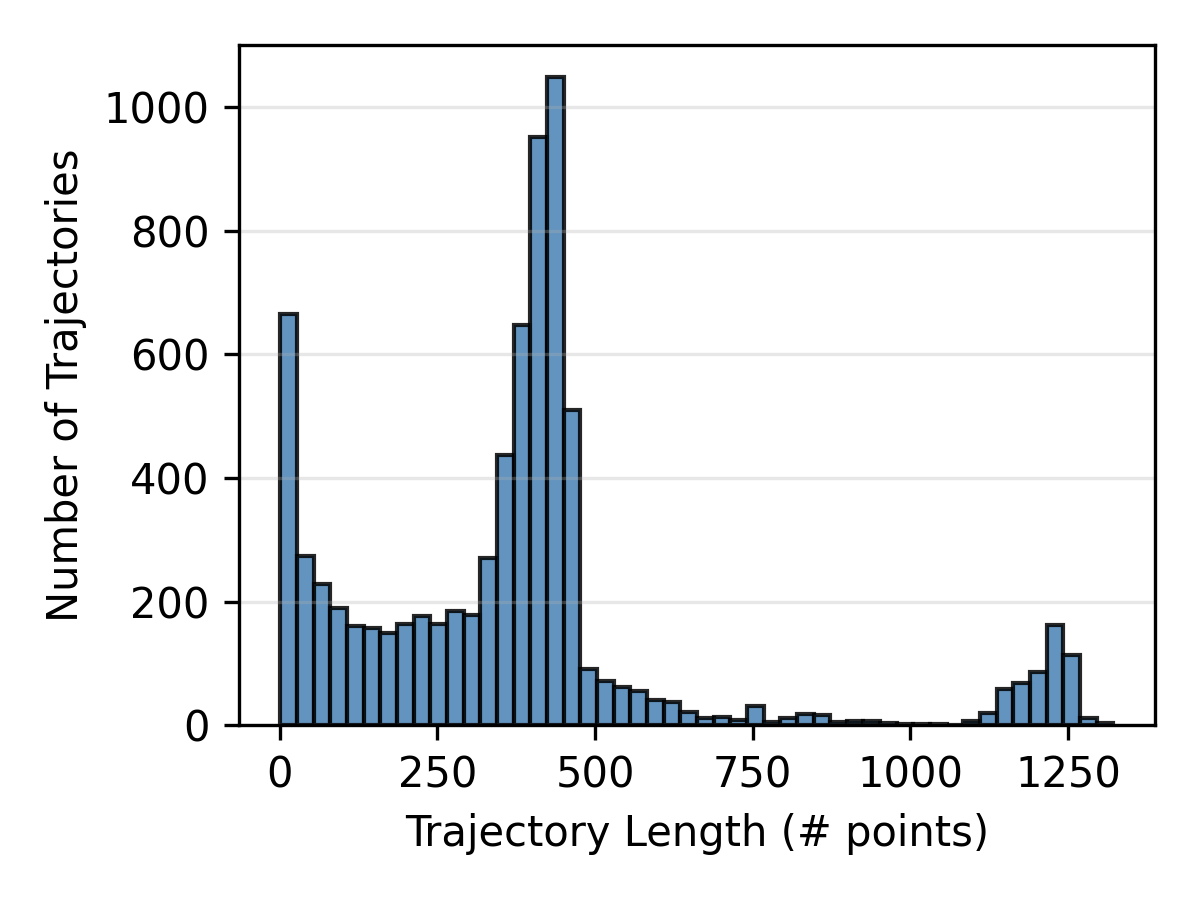}%
    \label{fig:tampa_hist}}

  \caption{Histogram of the Vessel trajectory length.}
  \label{fig:two-histogram}
  
\end{figure}

Additionally, Figure~\ref{fig:L_sensitivity} shows the performance variation with observation window $L\in \{40, 60, 80, 100\}$. It is observed that model's prediction performance is improved with increasing $L$. However, with the increasing $L$, the minimum required trajectory length $(L+T)$ is also increased. Additionally, the number of trajectories with smaller length is considerable in the histogram plot, shown in Figure~\ref{fig:two-histogram}. Considering this trade-off, and following the criteria of the M3 data challenge, we set the observation and prediction lengths to 60 and 120, respectively. \par
Moreover, the same window length ($L=60, T=120$) corresponds to different physical time spans (approximately 3 hours for AIS vs. 60 seconds for video(180 points)). However, our model is designed to be scale-agnostic by focusing on the relative temporal dynamics within a sequence. Since the model is trained and evaluated independently for each modality, it learns the “behavioral physics” appropriate to each domain, such as long-term navigational trends in AIS and short-term kinematic maneuvers in video. Additionally, the coverage area within an image is very small compared to that of the AIS data. Moreover, the vessel's trajectory across the river is very small compared to its movement along the river. Therefore, it is essential to use a sufficiently small sampling period to capture sufficient motion in the VVR data while rejecting noise arising from the high-frequency raw sampling rate (30 FPS).

\begin{table*}[!htb]
\centering
\scriptsize
\setlength{\tabcolsep}{5pt}
\caption{Absolute values of six metrics for each ablation case, corresponding to the relative percentage changes reported in Table~\ref{tab:ablation}. Symbol $\spadesuit$ in the Head column refers to our proposed head layer design, while symbol $\diamondsuit$ refers to the regular head layer design for time series forecasting models. The "$D$" in the RevIN column refers only to denormalization, whereas "$ND$" refers to both normalization and denormalization. RTA refers to the Residual Trajectory Adjuster module.}
\label{tab:ablation_abs_value}
\begin{tabular}{@{}ccccccc|cccccc@{}}
\toprule
\rotatebox{90}{Case} & \rotatebox{90}{RTA} & \rotatebox{90}{Emb. Mx} & \rotatebox{90}{Tm. Mx} & \rotatebox{90}{Head} & \rotatebox{90}{Padding} & \rotatebox{90}{RevIN} & \rotatebox{90}{MSE} & \rotatebox{90}{MAE} & \rotatebox{90}{SMAPE} & \rotatebox{90}{FDE} & \rotatebox{90}{FD} & \rotatebox{90}{AED} \\ \midrule
1 & \checkmark & \checkmark & \checkmark & $\spadesuit$ & \checkmark & $D$ & 3.86E-06 & 1.45E-03 & 4.08E-03 & 1.64E-03 & 3.72E-03 & 1.16E-03 \\
2 & $\times$ & \checkmark & \checkmark & $\spadesuit$ & \checkmark & $D$ & 4.41E-06 & 1.52E-03 & 4.26E-03 & 1.75E-03 & 3.79E-03 & 1.22E-03 \\
3 & \checkmark & \checkmark & $\times$ & $\spadesuit$ & \checkmark & $D$ & 6.10E-03 & 6.77E-02 & 1.64E-01 & 8.88E-02 & 9.17E-02 & 5.40E-02 \\
4 & \checkmark & $\times$ & \checkmark & $\spadesuit$ & \checkmark & $D$ & 1.63E-05 & 3.35E-03 & 9.79E-03 & 4.31E-03 & 6.96E-03 & 2.67E-03 \\
5 & \checkmark & \checkmark & \checkmark & $\spadesuit$ & \checkmark & $ND$ & 1.28E-05 & 2.42E-03 & 6.79E-03 & 3.25E-03 & 5.35E-03 & 1.95E-03 \\
6 & \checkmark & \checkmark & \checkmark & $\spadesuit$ & \checkmark & - & 5.37E-06 & 1.78E-03 & 5.13E-03 & 2.08E-03 & 4.24E-03 & 1.42E-03 \\
7 & \checkmark & \checkmark & \checkmark & $\spadesuit$ & $\times$ & $D$ & 1.19E-05 & 2.89E-03 & 8.09E-03 & 3.74E-03 & 6.07E-03 & 2.32E-03 \\
8 & \checkmark & \checkmark & \checkmark & $\diamondsuit$ & \checkmark & $D$ & 2.30E-05 & 4.06E-03 & 1.19E-02 & 5.50E-03 & 7.97E-03 & 3.26E-03 \\ \bottomrule
\end{tabular}
\end{table*}

\section{Semantic Consistency of Coordinate Systems}
\label{semantic_consistency_of_coordinate_systems}
A key challenge in multi-modal trajectory prediction is the semantic difference between coordinate systems. In this study, AIS data utilizes geographic spherical coordinates (Latitude/Longitude), while the VVR dataset uses normalized image-plane coordinates ($xy$). We justify the unified treatment of these data types based on three factors. First, the VVR camera remains stationary with a fixed orientation, ensuring that image-plane coordinates provide a stable and consistent spatial reference frame throughout the dataset. Second, the use of z-normalization (as detailed in Section~\ref{dataset_normalization}) maps both coordinate systems to a non-dimensional scale, allowing the Mix\&Fix-Net to focus on relative motion dynamics rather than absolute spatial metrics. Finally, the model is trained and evaluated independently on AIS and non-AIS datasets. This allows the model to adapt to different multi-modal datasets. The high performance across both modalities confirms that the architecture is robust and coordinate-agnostic.


\end{document}